\documentclass[11pt,letterpaper]{article}

\usepackage[letterpaper,margin=1in]{geometry}
\usepackage{graphicx}
\usepackage{amsmath,amssymb}
\usepackage{booktabs}
\usepackage[numbers,square,sort&compress]{natbib}
\usepackage{float}
\usepackage{subcaption}
\usepackage{xcolor}
\usepackage{caption}
\usepackage{microtype,multirow}
\usepackage{titling}

\usepackage{amsmath,amsfonts,bm}

\def\eqref#1{equation~\ref{#1}}
\def\1{\bm{1}}

\DeclareMathAlphabet{\mathsfit}{\encodingdefault}{\sfdefault}{m}{sl}
\SetMathAlphabet{\mathsfit}{bold}{\encodingdefault}{\sfdefault}{bx}{n}

\definecolor{paperblue}{rgb}{0.21,0.49,0.74}
\usepackage[breaklinks,colorlinks,allcolors=paperblue]{hyperref}

\newcommand{\bench}{\textsc{MultiWorldBench}}

\pretitle{\centering\LARGE}
\posttitle{\par\vspace{0.4em}}

\preauthor{\centering\large\tabular[t]{c}}
\postauthor{\endtabular\par}

\predate{}
\postdate{}

\title{%
\bench: Do Independently Controlled Views\\
Describe One Shared World?
}

\author{%
Zhangbo Xu\textsuperscript{1} \quad
Ruoxi Zhang\textsuperscript{1} \quad
Rui Hu\textsuperscript{1}\thanks{Correspondence: Rui Hu
\textless\href{mailto:hurui1983@gmail.com}{hurui1983@gmail.com}\textgreater}
\quad
Yisong Wang\textsuperscript{1} \quad
Xitong Li\textsuperscript{2,3}\thanks{Email: Xitong Li
\textless\href{mailto:x.li1@imperial.ac.uk}{x.li1@imperial.ac.uk}\textgreater}
\quad
Jingdong Wang\textsuperscript{4}\\[0.5em]
\textsuperscript{1}Zhiying Guangnian Technology Co., Ltd.\\
\textsuperscript{2}Imperial College London
\quad
\textsuperscript{3}HEC Paris
\quad
\textsuperscript{4}Independent Researcher
}

\date{}

\begin{document}

\maketitle
\vspace{-1.5em}

\begingroup
\setlength{\intextsep}{6pt}

\begin{figure}[H]
    \centering
    \includegraphics[
        width=0.85\linewidth
    ]{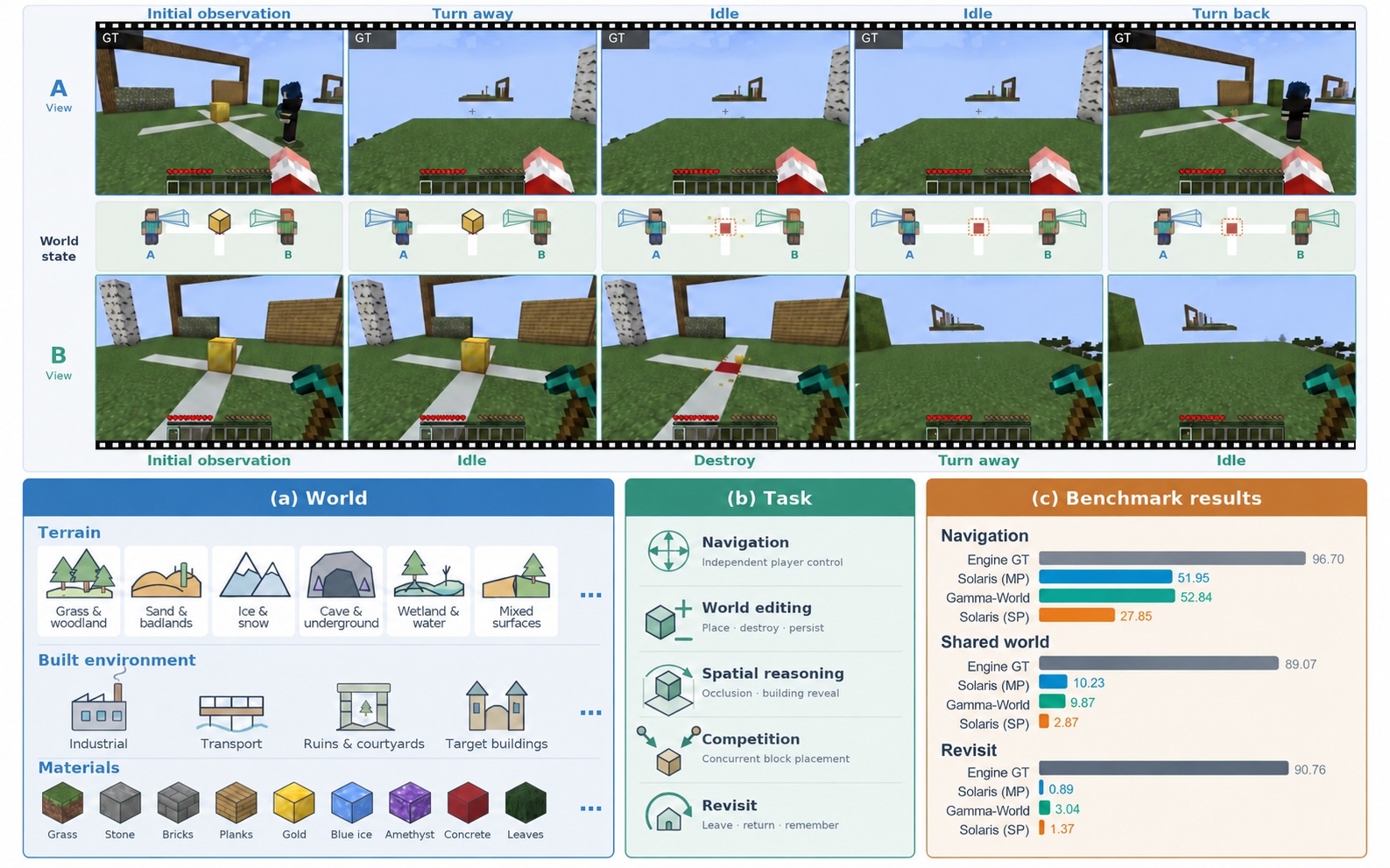}

    \captionsetup{font=footnotesize,skip=4pt}
    \caption{
        \textbf{Overview of \bench{}.}
        \textbf{Top:} A turns away while B destroys the gold block;
        when A returns, the edit persists. The middle row depicts
        the evolving world state and viewing directions.
        \textbf{Bottom:} (a) terrain, built-environment, and material
        diversity; (b) five task families; and (c) measured
        control-isolation, state-synchronization, and
        state-persistence scores from Table~\ref{tab:scorecard}
        (0--100; higher is better).
    }
    \label{fig:overview}
\end{figure}

\endgroup

\noindent
\begin{minipage}{\linewidth}
\begin{abstract}
Multiplayer world models must ensure that independently controlled views
remain consistent with one shared and persistent world. We introduce
\bench, a diagnostic Minecraft benchmark containing 495 case configurations
across seven task suites and ten capabilities, including independent
control, cross-view motion, shared-state synchronization, persistence,
structural reasoning, concurrent interaction, and delayed revisit. We
evaluate two multiplayer world models, Solaris (MP) and Gamma-World,
together with a single-player generation variant, Solaris (SP), using
Engine GT as a reference. Gamma-World achieves the highest ten-capability
average among the generated systems at 21.39, followed by Solaris (MP) at
20.88 and Solaris (SP) at 10.06, while Engine GT reaches 91.69. Gamma-World
performs better on several control, shared-state, and revisit capabilities,
whereas Solaris (MP) leads in cross-view motion and race-condition
consistency. Nevertheless, all generated systems score at most 8.00 on
state persistence and 1.33 on structural consistency, and none succeeds in
spatial reasoning or building-identity preservation. Human pairwise evaluation yields the same overall system ranking and shows strong alignment with the automatic evaluation, with a mean dimension-level Spearman correlation of 0.95. These results show that current multiplayer world models still
exhibit a substantial gap from Engine GT in maintaining a coherent shared
world.
\end{abstract}
\end{minipage}
\par

\section{Introduction}
\label{sec:introduction}

Interactive world models have developed rapidly from learned latent
simulators into action-conditioned video generators that support navigation
and intervention~\citep{ha2018worldmodels,bruce2024genie,alonso2024diamond}.
The emerging multiplayer setting introduces a more demanding objective:
several independently controlled views must describe one persistent world.
A valid system must respond to each player's own actions, isolate that
player from unrelated controls, synchronize changes that belong to the
shared state, and preserve entities when they leave and later re-enter
view. Recent multiplayer world models indicate that this setting is
becoming technically feasible~\citep{savva2026solaris,liu2026gamma,hu2026multiplayer}.

Minecraft provides a particularly useful testbed because it offers a stable
control interface, scalable trajectory collection, discrete materials, and
world changes that can be checked at known grid locations. Recent work has
consequently introduced synchronized multiplayer data and model
architectures in Minecraft~\citep{savva2026solaris,liu2026gamma}. Yet
evaluation has not advanced at the same pace. Solaris includes an
evaluation suite for multiplayer generation~\citep{savva2026solaris}, but
current protocols provide limited variation in scene geometry, coarse
summaries of distinct failure modes, and a narrow range of evaluation
methods. As a result, they cannot fully distinguish whether a model follows
each player's controls, leaks one player's actions into another view,
reproduces the same physical motion across viewpoints, synchronizes an
intervention, or preserves the identity of a revisited object. Per-view
visual quality is also insufficient: two individually plausible videos may
disagree on the number, material, motion, or location of objects.
Table~\ref{tab:benchmark-comparison} positions these differences against
representative video and interactive-world benchmarks.

Figure~\ref{fig:overview} summarizes the world configurations and task
families in \bench{}. The persistence case illustrates how an edit made by
one player must remain visible when another player later returns to the
edited region. In a shared-state case, both players observe a colored cross
marking one grid location. One or both players place or destroy a block at
its center while their initial visibility is controlled. A valid rollout
must execute the requested action, show a consistent block count and target
location in both views, and, for placement, preserve the intended material.
In a more challenging building-reveal case, player Bravo initially sees
both player Alpha and a building, whereas Alpha sees an occluder. After
Alpha removes the occluder, Alpha should observe the same building that
remains visible to Bravo. These cases turn the abstract notion of a
``shared world'' into falsifiable cross-view observations.

To close this gap, we introduce \bench{}, a benchmark organized around three
requirements. The first is \emph{independent control}, which asks whether
each player responds to its assigned actions without interference from the
other player and whether the resulting motion is compatible across views.
The second is \emph{world consistency}, which asks whether both views
support the same shared state. The third is \emph{revisit consistency},
which asks whether visual appearance and object identity are preserved after
a target leaves both views and is later observed again by another player.
Accordingly, the benchmark contains three navigation metrics, five
shared-world metrics, and two revisit metrics. Specifically, it evaluates
control isolation, coordination fairness, cross-view motion, state
synchronization, state persistence, structural consistency, spatial
reasoning, race-condition consistency, style consistency, and
building-identity consistency.

\bench{} is designed around three principles. First, it orders cases
according to the visual information shared between players. Players may be
mutually visible, visible in only one direction, or mutually invisible.
This progression reduces direct cross-view evidence and provides a
candidate easy-to-hard ordering. The cases retain these visibility labels
for stratified evaluation, although the aggregate results reported here do
not by themselves establish such a difficulty ordering. Second, the
benchmark covers the major Minecraft action families supported by current
multiplayer video models. Navigation includes forward and backward
movement, lateral movement, rotation, idle controls, and cross-view motion
responses. The shared-state suite combines placement and destruction with
one or two active players. Concurrent placement, occluder removal, and
delayed cross-player revisit extend this taxonomy to conflicting actions,
structural reveal, and long-term world persistence. Third, \bench{} uses
evaluation methods matched to the failure being tested. Estimated camera
trajectories measure player-specific control responses; action-grounded
visual judgments assess whether an actor's movement produces the expected
relative displacement in both views; binary visual judgments evaluate
shared block states and building identity; segmentation and viewpoint
estimation measure structural and spatial consistency; and color
distributions measure style consistency after delayed revisit. This
combination produces an interpretable capability profile instead of
collapsing heterogeneous behaviors into a single video-quality score.

We evaluate two multiplayer world models, Solaris (MP)~\citep{savva2026solaris}
and Gamma-World~\citep{liu2026gamma}, together with a single-player
generation variant, Solaris (SP), on 495 benchmark cases, using Engine GT
as a reference. Three principal findings emerge.
\textbf{First, multiplayer world models better coordinate information
across independently controlled views, but remain far from the Engine-GT
reference.}
Solaris (MP) and Gamma-World substantially outperform Solaris (SP) in
state synchronization, cross-view motion, and race-condition consistency,
all of which require actions or state changes to be coordinated across
views. This comparison suggests that multiplayer-oriented generation
provides a better foundation for maintaining cross-view correspondence.
Nevertheless, both multiplayer models still exhibit control leakage,
inaccurate motion magnitude, and incompatible responses across viewpoints,
leaving a substantial gap from Engine GT.
\textbf{Second, errors in basic actions and instruction execution propagate
to more complex shared-world tasks.}
All three generated systems may move too little or too far, continue moving
after a command ends, fail to execute a block operation, or apply the wrong
operation, material, or location. Because synchronization, persistence,
building reveal, spatial reasoning, and concurrent interaction depend on
correct movement and a valid initial world state, these basic failures can
invalidate a later task before its target capability is meaningfully
tested.
\textbf{Third, current systems demonstrate little reliable world memory.}
The generated systems frequently lose edited states, scene structure, or
object identity after a region leaves view, even when coarse visual style
remains similar. Human preferences reproduce the same overall system
ordering and show strong rank agreement with the automatic evaluation.
These findings are descriptive results under the recorded evaluation
protocols rather than claims of statistical significance. Overall, current
multiplayer world models still exhibit a substantial gap from Engine GT in
maintaining a coherent and persistent shared world.

Our contributions are threefold:
(1) a visibility-conditioned benchmark that varies cross-player evidence
from mutual visibility to mutual invisibility and supports stratified
evaluation;
(2) a 495-case, seven-suite action taxonomy spanning navigation,
cross-view motion, synchronized and persistent interventions, concurrent
actions, building reveal, and delayed revisit; and
(3) a ten-capability evaluation system combining geometric,
trajectory-based, VLM-based, segmentation-based, viewpoint-based, and
distributional measurements, together with a matched comparison of two
multiplayer world models, one single-player generation variant, an
Engine-GT reference, and human-preference validation.

\section{Related Work}
\label{sec:related-work}

\paragraph{World models and interactive generation.}
World models learn predictive dynamics to support action and
planning~\citep{ha2018worldmodels}. Genie and DIAMOND demonstrate
complementary approaches to interactive visual
environments~\citep{bruce2024genie,alonso2024diamond}.
Minecraft-based systems such as MineWorld further explore
action-conditioned generation in interactive
environments~\citep{guo2025mineworld}. These single-player settings
evaluate the evolution of an individual observation stream, but do
not directly test whether independently controlled player views
remain consistent with one shared world.

\paragraph{Video and interactive benchmarks.}
VBench and EvalCrafter evaluate perceptual, semantic, motion, and
temporal video quality~\citep{huang2024vbench,liu2024evalcrafter},
while T2V-CompBench evaluates compositional generation, including
motion and spatial relationships~\citep{sun2025t2vcompbench}.
Interactive-world benchmarks extend evaluation to action-conditioned
and multi-turn generation. WBench evaluates multi-turn
interaction~\citep{ying2026wbench}, MIND examines action control
and revisit memory~\citep{ye2026mind}, and WorldMark evaluates
action dynamics and world memory~\citep{xu2026worldmark}.
GameWorld also evaluates interactive generation in game
environments~\citep{zhang2025matrixgame}.
These evaluations address capabilities relevant to interactive world
modeling, but do not directly assess whether changes induced by
independently controlled players remain consistent across their
respective views.

\paragraph{Multiplayer world-model evaluations.}
Solaris introduces task-specific evaluations of movement, grounding,
memory, building, and view consistency for multiplayer Minecraft
generation~\citep{savva2026solaris}. Gamma-World reports multiplayer
generation results using distributional and perceptual metrics,
including FID and FVD~\citep{liu2026gamma}.
The evaluation accompanying ME-World measures camera control,
self-action control, and other-agent action control, together with
shared-environment consistency, interaction-induced state-update
consistency, and agent-identity preservation on real and synthetic
multi-agent data~\citep{chung2026meworld}.

These evaluations provide important foundations for measuring
multiplayer world models. However, they do not jointly provide the
controlled visibility stratification and capability-specific
measurements needed to distinguish independent control, role-level
coordination, cross-view motion, shared-state propagation,
concurrent-edit resolution, and delayed cross-player identity
preservation.

Table~\ref{tab:benchmark-comparison} summarizes the coverage of
representative benchmarks and model-specific evaluation suites under
the multiplayer capability definitions used in this work.
\bench{} brings these capabilities into a unified evaluation framework,
combining controlled visibility, independent player control,
shared-state interventions, and delayed cross-player revisits.
It contains 495 case configurations across seven task suites and
provides ten metrics organized into Navigation, Shared World,
and Revisit.

\begin{table}[H]
\centering
\small
\setlength{\tabcolsep}{1.5pt}
\renewcommand{\arraystretch}{1.05}

\caption{
Comparison under our multiplayer capability definitions.
\checkmark: direct evaluation;
\textit{L}: limited or indirect coverage;
$\times$: no corresponding evaluation identified.
Scale follows each source's counting unit;
-- indicates an unreported count.
}
\label{tab:benchmark-comparison}

\begin{tabular*}{\textwidth}{
@{\extracolsep{\fill}}
l
ccc
ccc
ccccc
cc
c
@{}
}
\toprule
& \multicolumn{3}{c}{Paired visibility}
& \multicolumn{3}{c}{Navigation}
& \multicolumn{5}{c}{Shared world}
& \multicolumn{2}{c}{Revisit}
& Scale \\
\cmidrule(lr){2-4}
\cmidrule(lr){5-7}
\cmidrule(lr){8-12}
\cmidrule(lr){13-14}
\cmidrule(lr){15-15}

Benchmark
& $A\leftrightarrow B$
& $A\rightarrow B$
& $A\mid B$
& Iso.
& Fair.
& Mot.
& Sync.
& Pers.
& Str.
& Spa.
& Race
& Style
& ID
& $N$ \\
\midrule

\multicolumn{15}{@{}l}{
\textit{Video-generation benchmarks}
} \\[1pt]

VBench~\citep{huang2024vbench}
& $\times$ & $\times$ & $\times$
& $\times$ & $\times$ & $\times$
& $\times$ & $\times$ & $\times$ & $\times$ & $\times$
& $\times$ & $\times$
& 946 \\

\cmidrule(lr){1-15}

\multicolumn{15}{@{}l}{
\textit{Interactive world-model benchmarks and evaluations}
} \\[1pt]

WBench~\citep{ying2026wbench}
& $\times$ & $\times$ & $\times$
& $\times$ & $\times$ & $\times$
& $\times$ & $\times$ & $\times$ & $\times$ & $\times$
& \textit{L} & \textit{L}
& 289 \\

MIND~\citep{ye2026mind}
& $\times$ & $\times$ & $\times$
& \textit{L} & $\times$ & \textit{L}
& $\times$ & \textit{L} & $\times$ & $\times$ & $\times$
& \textit{L} & \textit{L}
& 250 \\

WorldMark~\citep{xu2026worldmark}
& $\times$ & $\times$ & $\times$
& \textit{L} & $\times$ & \textit{L}
& $\times$ & \textit{L} & \textit{L} & \textit{L} & $\times$
& \textit{L} & \textit{L}
& 500 \\

MineWorld Eval.~\citep{guo2025mineworld}
& $\times$ & $\times$ & $\times$
& $\times$ & $\times$ & $\times$
& $\times$ & $\times$ & $\times$ & $\times$ & $\times$
& $\times$ & $\times$
& 1,000 \\

\cmidrule(lr){1-15}

\multicolumn{15}{@{}l}{
\textit{Multiplayer world-model evaluations}
} \\[1pt]

Gamma-World~\citep{liu2026gamma}
& $\times$ & $\times$ & $\times$
& $\times$ & $\times$ & $\times$
& $\times$ & $\times$ & $\times$ & $\times$ & $\times$
& $\times$ & $\times$
& -- \\

Solaris Eval.~\citep{savva2026solaris}
& \checkmark & \checkmark & $\times$
& \textit{L} & $\times$ & \checkmark
& \checkmark & \textit{L} & \checkmark & \textit{L} & $\times$
& $\times$ & $\times$
& -- \\

ME-World Eval.~\citep{chung2026meworld}
& \textit{L} & $\times$ & $\times$
& \textit{L} & $\times$ & \textit{L}
& \checkmark & \textit{L} & \textit{L} & \textit{L} & $\times$
& \textit{L} & \textit{L}
& 200 \\

\addlinespace[2pt]

\textbf{\bench{}}
& \textbf{\checkmark} & \textbf{\checkmark} & \textbf{\checkmark}
& \textbf{\checkmark} & \textbf{\checkmark} & \textbf{\checkmark}
& \textbf{\checkmark} & \textbf{\checkmark} & \textbf{\checkmark}
& \textbf{\checkmark} & \textbf{\checkmark}
& \textbf{\checkmark} & \textbf{\checkmark}
& 495 \\

\bottomrule
\end{tabular*}
\end{table}

\section{Benchmark Design}
\label{sec:design}
\subsection{Paired Observations and Independent Actions}
Let \(a_t^i\) and \(x_t^i\) denote the native action and observation of player \(i \in \{A,B\}\), respectively. The model generates
\begin{equation}
p_{\theta}!\left(
x_{t+1}^{A}, x_{t+1}^{B}
\mid
x_{\leq t}^{A}, x_{\leq t}^{B},
a_{\leq t}^{A}, a_{\leq t}^{B}
\right).
\end{equation}
The two views should be consistent with a single shared world state, while each player's camera remains driven by its own controls. The benchmark evaluates the observable consequences of this requirement rather than assuming access to the model's internal state.
Engine recordings execute native movement, camera, placement, and destruction actions. Scene initialization is separate from the recorded trajectory. Case files specify role poses, target geometry, action intervals, expected changes, and evaluation frames. When a rollout is not collected in the engine, its supplied action schedule is treated as planned conditioning rather than as an executed ground-truth action log.

\subsection{Dataset Statistics}
\label{sec:dataset-statistics}

\begin{figure}[H]
    \centering
    \includegraphics[width=\textwidth]
    {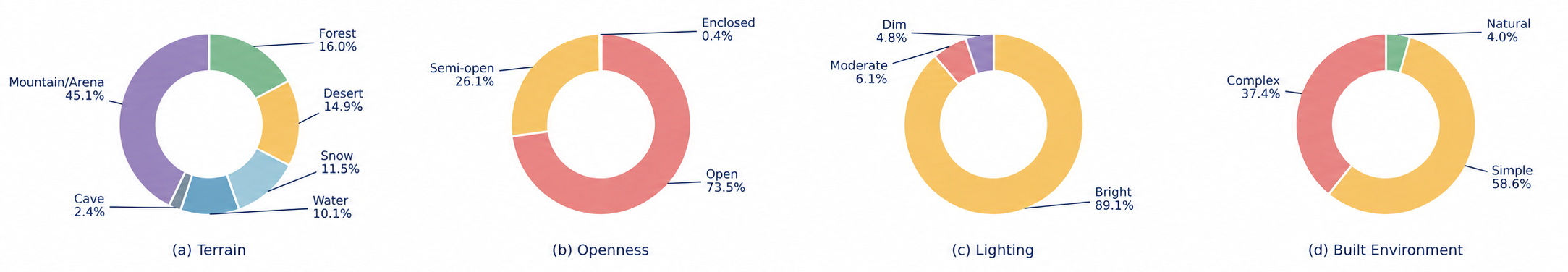}
    \caption{
        \textbf{Dataset composition in \textsc{MultiWorldBench}.}
        Distribution of 495 cases by terrain, spatial openness, lighting,
        and built-environment complexity.
    }
    \label{fig:dataset-diversity-composition}
\end{figure}

\textsc{MultiWorldBench} comprises 495 evaluation cases spanning diverse
Minecraft environments. We annotate each case along four complementary
dimensions: terrain, spatial openness, lighting, and built-environment
complexity, as summarized in
Fig.~\ref{fig:dataset-diversity-composition}. The 20 cross-view-motion
cases use a controlled marker arena and are categorized as
mountain/arena, open, bright, and simple. Representative frames are
provided in Appendix~\ref{app:dataset-gallery}.

\textbf{Terrain Diversity.}
The dataset contains six terrain groups
(Fig.~\ref{fig:dataset-diversity-composition}(a)). The merged
mountain/arena group is the largest (45.1\%), followed by forest
(16.0\%), desert (14.9\%), snow (11.5\%), water (10.1\%), and cave
environments (2.4\%). These groups cover natural biomes, underground
spaces, and controlled or constructed environments.

\textbf{Spatial Openness and Lighting.}
Most cases occur in open environments (73.5\%), followed by semi-open
(26.1\%) and enclosed environments (0.4\%), as shown in
Fig.~\ref{fig:dataset-diversity-composition}(b). Lighting is predominantly
bright (89.1\%), with moderate (6.1\%) and dim (4.8\%) cases providing
more challenging visibility conditions
(Fig.~\ref{fig:dataset-diversity-composition}(c)).

\textbf{Built-Environment Complexity.}
As shown in Fig.~\ref{fig:dataset-diversity-composition}(d), simple built
environments constitute 58.6\% of the dataset, complex built environments
account for 37.4\%, and natural environments account for 4.0\%. These
groups range from scenes without salient constructed structures to
environments containing sparse constructions or dense architectural
layouts.

\subsection{Evaluation case construction}

Each case specifies \textbf{1)~scene geometry and initial visibility},
\textbf{2)~player roles and target locations}, and
\textbf{3)~an action schedule with expected changes and evaluation frames}.
The benchmark contains 495 case configurations across seven task suites
supporting ten capabilities. Initial visibility is mutual, one-way,
or absent and describes only the starting configuration. Geometry-rich
scenes provide stronger camera-tracking constraints than flat environments,
although they cannot eliminate reconstruction error. For block-intervention
tasks, colored crosses mark target locations, expected block counts are
specified explicitly, and dropped items are excluded from full-block counts.

\noindent\textbf{1) Matched navigation.}
Six scenes, three visibility conditions, and ten schedules per
scene--visibility combination yield 180 cases. Each case tests both
players as the target, repeating the target player's action while
the other player is idle, performs the same action, or performs a
different action. The schedules include forward and backward movement,
left and right strafing, yaw and pitch rotation, combined translation
and rotation, and idle controls.
For readability, illustrative controls are denoted as \texttt{W} (forward),
\texttt{S} (backward), \texttt{A} (left strafe),
$\leftarrow$ (leftward camera rotation), and idle.
Representative paired views for the three visibility conditions and
examples of the native navigation controls are shown in
Fig.~\ref{fig:visibility-navigation-controls}. Scene and visibility
assignments are balanced across action conditions. These cases support
both control-isolation and coordination-fairness evaluation.

\begin{center}
\begin{minipage}{\linewidth}
    \centering
    \includegraphics[
        width=\linewidth,
        trim=0 0 0 0,
        clip
    ]{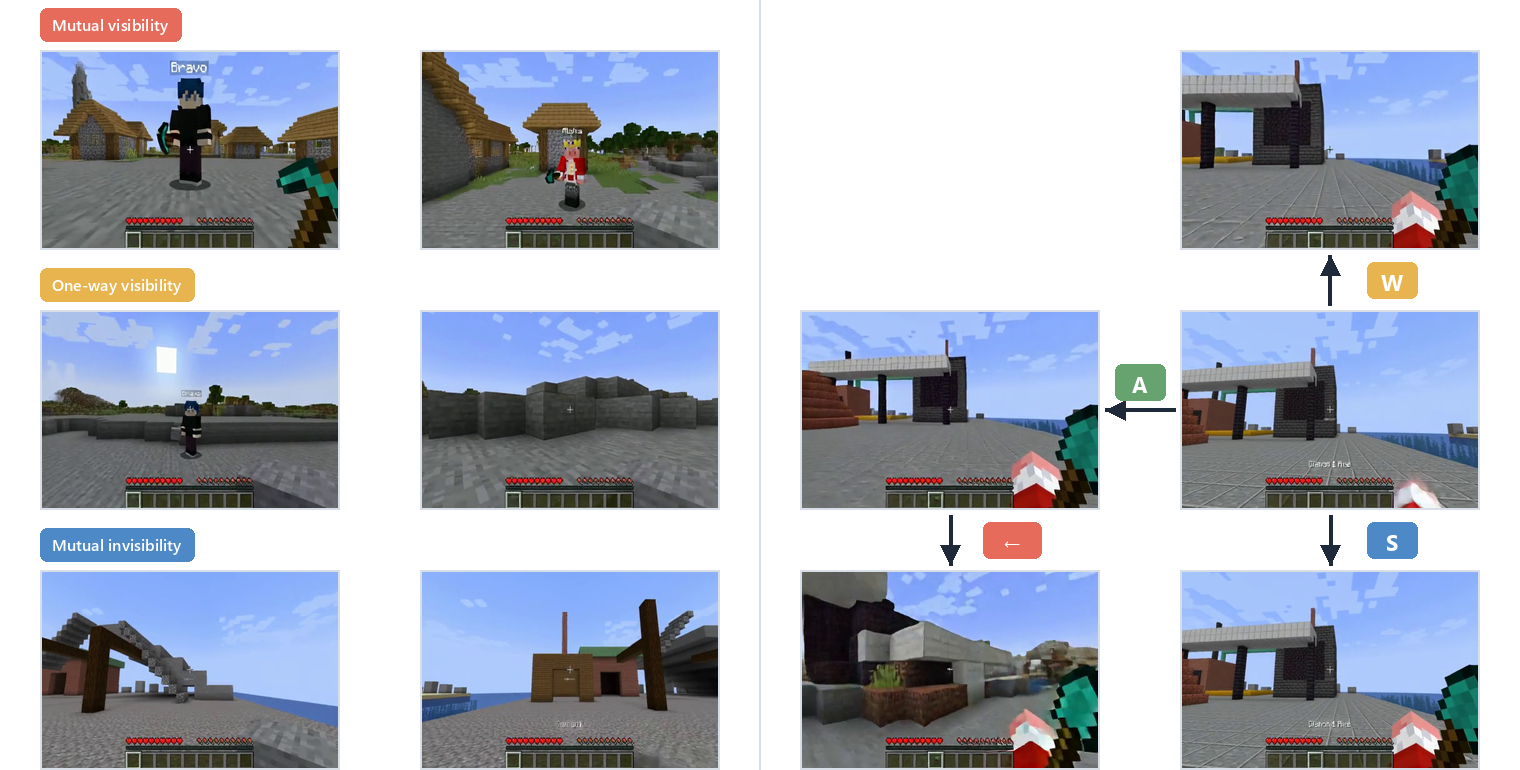}

    \captionof{figure}{
        \textbf{Visibility configurations and navigation controls.}
        The left panel presents paired agent views under mutual visibility,
        one-way visibility, and mutual invisibility. The right panel
        illustrates the native navigation controls, including forward and
        backward movement, left strafing, and leftward camera rotation.
        Frames within each example are sampled from the same continuous
        video.
    }
    \label{fig:visibility-navigation-controls}
\end{minipage}
\end{center}

\noindent\textbf{2) Cross-view motion.}
Twenty cases evaluate whether the actor's movement is represented
consistently across the two player views. The actor remains visible
to the other player, enabling comparison between the movement
indicated by the actor's own view and that observed from the
partner's view.

\noindent\textbf{3) Shared-state interaction.}
Five scenes, three visibility conditions, four interaction types,
and two variants yield 120 cases. The interaction types combine
one or two active players with placement or destruction, producing
60 cases for each operation. The suite tests whether the intended
edit succeeds and whether the resulting world state appears
consistently across the two player views. Representative
before-and-after examples are shown in
Fig.~\ref{fig:shared-state-edits}.

\begin{center}
\begin{minipage}{\linewidth}
    \centering
    \includegraphics[
        width=\linewidth,
        trim=0 0 0 0,
        clip
    ]{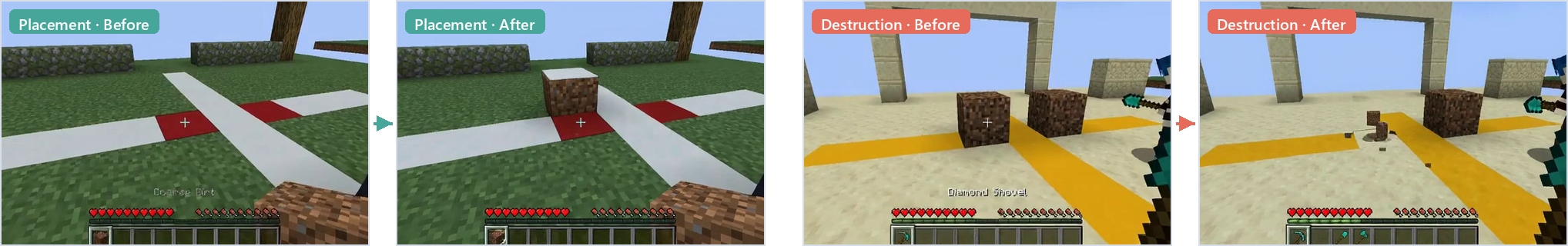}

    \captionof{figure}{
        \textbf{Representative shared-state interactions.}
        From left to right: the marked target before placement,
        the resulting placed block, the scene before destruction,
        and the scene after destruction. Teal denotes placement
        and coral denotes destruction. Each before--after pair
        is sampled from the same continuous rollout.
    }
    \label{fig:shared-state-edits}
\end{minipage}
\end{center}

\noindent\textbf{4) Shared-state persistence.}
Five environments, five block materials, and two operations yield
50 cases. One player places or destroys a block at the center of
a visible cross, after which the other player returns to the
operation region. A rollout is successful only when the expected
edit is visible to the operator after the action and remains
visible to the returning observer.

\noindent\textbf{5) Building reveal.}
In 75 cases, A's initial view is occluded while B can see the target
building. A then removes the occluder to reveal the building,
as illustrated in the first two panels of
Fig.~\ref{fig:reveal-revisit}. Successful revelation is a shared
prerequisite for structural-consistency and spatial-reasoning
evaluation, so both metrics use the same cases.

\noindent\textbf{6) Concurrent placement.}
In 30 cases, both players attempt to place different candidate
materials at the same location and time using a single placement
event. The two views are compared for agreement on block count,
block type, and whether the result matches one of the intended
materials.

\noindent\textbf{7) Delayed cross-player revisit.}
Ten scene configurations and their role-swapped counterparts yield
20 directional cases. The first observer sees the building and
turns away; after the building remains outside both views for
a 70-frame idle interval (3.5 seconds at 20~Hz), the second observer
turns toward it. The evidence pair compares frames 20 and 155,
while role swapping exchanges the early and late observers.
Unlike a same-camera return, this protocol tests appearance and
identity persistence across both time and player identity.
Representative occlusion, reveal, and cross-player revisit views
are summarized in Fig.~\ref{fig:reveal-revisit}.

\begin{center}
\begin{minipage}{\linewidth}
    \centering
    \includegraphics[
        width=\linewidth,
        trim=0 0 0 0,
        clip
    ]{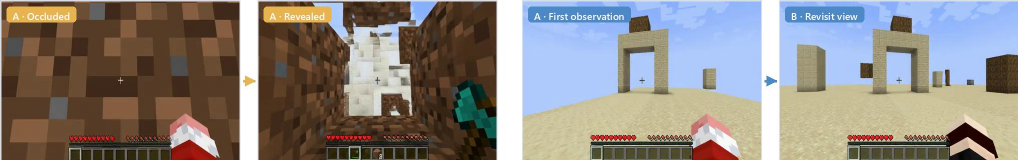}

    \captionof{figure}{
        \textbf{Building reveal and cross-player revisit.}
        The first two panels show A's initially occluded view
        and the target exposed after removal of the occluding block.
        The last two panels show the same target from the two
        observer roles used in the role-swapped revisit protocol.
        Orange denotes the occlusion--reveal transition, while
        blue denotes the cross-player observation pair.
    }
    \label{fig:reveal-revisit}
\end{minipage}
\end{center}

\begin{center}
\begin{minipage}{\linewidth}
    \centering
    \small
    \captionof{table}{
        Evaluation tasks and case support.
        The seven task suites contain 495 case configurations.
        A task suite may support multiple metrics; cases shared
        by multiple metrics are counted only once.
    }
    \label{tab:cohorts}

    \begin{tabular}{@{}p{.27\linewidth}rp{.52\linewidth}@{}}
        \toprule
        Task & Cases & Evaluation target \\
        \midrule

        Matched navigation
        & 180
        & Control isolation and coordination fairness \\

        Cross-view motion
        & 20
        & Consistency between actor-view and partner-view motion \\

        Shared-state interaction
        & 120
        & Placement/destruction agreement across views \\

        Shared-state persistence
        & 50
        & Operator and returning-observer change agreement \\

        Building reveal
        & 75
        & Structural identity, revealed area, and relative viewpoint \\

        Concurrent placement
        & 30
        & Count, type, and intended-material agreement \\

        Delayed cross-player revisit
        & 20
        & Color-distribution and building-ID persistence \\

        \midrule
        \textbf{Total}
        & \textbf{495}
        & \textbf{Seven task suites supporting ten capabilities} \\
        \bottomrule
    \end{tabular}
\end{minipage}
\end{center}

\section{Evaluation}

Scores retain their native scales in the stored artifacts. We display all results on $[0,100]$ for readability, multiplying binary or normalized scores by 100. This display conversion does not make heterogeneous metrics interchangeable.

\subsection{Navigation}

\paragraph{Response features.}
The navigation evaluator estimates camera poses with MCV using known-map 2D--3D correspondences and PnP. Videos are processed at 480p with camera intrinsics transformed consistently with image resizing and cropping; point replenishment, robust tracking, and backward recovery improve pose coverage. Translation is expressed in the map coordinate scale, without additional scaling to GT motion. For each action interval, the translational response comprises planar endpoint displacement, planar path length, and RMS deviation from the endpoint chord normalized by path length. Rotation is represented by a start-relative path on $\mathrm{SO}(3)$ and its accumulated angular travel, rather than endpoint angle alone.

For non-negative translational feature samples $X,Y$, let $\widehat{W}_1(X,Y)$ be the implemented empirical-quantile approximation to one-dimensional Wasserstein distance. Define
\begin{equation}
\label{eq:feature_similarity}
s_f(X,Y)
=
100\left[
1-
\min\left(
1,
\frac{
\widehat{W}_1(X,Y)
}{
\max\left(
\operatorname{mean}|X|
+
\operatorname{mean}|Y|,
\epsilon_f
\right)
}
\right)
\right].
\end{equation}

The denominator floors are 0.5 map units for endpoint displacement and path length, and 0.1 for normalized chord deviation; rotation comparisons use a $10^\circ$ floor. Similarity alone can reward equal inactivity, so commanded-motion response checks are applied as specified below. Insufficient pose evidence is reported separately from an observed control failure. The recorded navigation run requires at least three common valid action offsets across the four isolation repetitions and a contiguous jointly valid run of at least three samples for a fairness response. Optional bounded-gap interpolation uses linear translation and quaternion interpolation between reliable anchors within an action interval, with no extrapolation; it must be declared in the run configuration and was disabled in the stored 180-case run.

\paragraph{Control isolation.}
Comparisons are constructed within a case: the target repeats the same movement and camera command while the partner's command changes. Two repetitions provide reference responses. Let $P_{r,h,j}^{(k)}$ denote the start-relative response profile for target role $r$, channel $h\in\{\mathrm{translation},\mathrm{rotation}\}$, reference repetition $j\in\{1,2\}$, and profile sample $k$, and let $Q_{r,h,c}^{(k)}$ denote the response under partner condition $c$. Profiles preserve the recorded command progress and elapsed time. Define
\begin{equation}
\label{eq:control_isolation}
\begin{aligned}
e_{r,h,c}
&=
\frac12\sum_{j=1}^{2}
\frac{
K^{-1}\sum_{k=1}^{K}
d_h(P_{r,h,j}^{(k)},Q_{r,h,c}^{(k)})
}{
\max(L_h(P_{r,h,j}),\epsilon_h)
},\\
S_{\mathrm{isolation}}
&=
100\min_{r,h,c}
\left[1-\min(1,e_{r,h,c})\right].
\end{aligned}
\end{equation}

Thus, the case score is the worst-case similarity across target roles, response channels, and partner conditions. Here $d_{\mathrm{translation}}$ is Euclidean position distance, $d_{\mathrm{rotation}}$ is the geodesic rotation angle, and $L_h$ is the reference path length in the corresponding units. The floors are $\epsilon_{\mathrm{translation}}=0.5$ map units and $\epsilon_{\mathrm{rotation}}=10^\circ$. Both channels are evaluated for every action, including an uncommanded channel. Repeated-reference error is diagnostic only and is not subtracted from the treatment error.

A WBench-style repeated-trajectory control-consistency check using translation and rotation cnATE must reach the predefined threshold of $0.6$; otherwise isolation is unavailable. Only the target trajectory is required for its directional comparison. Static frames establish a noise baseline from the configured initial idle interval, or the final one-second idle interval when no initial interval is specified. With sufficient pose and noise evidence, a commanded reference response at or below the larger of the measured noise and response floor receives zero; fewer than three valid noise samples makes the comparison unavailable. The uncommanded channel uses the denominator floor without a motion-response gate, and treatment responses are not motion-gated. Thus passing consistency is an eligibility condition, not a correction that removes residual repeatability error.

\paragraph{Coordination fairness.}
Within each case, response samples from A and B are grouped by identical recorded movement and camera commands. Let $\mathcal H_c$ contain only the channels commanded by action class $c$. For translation, $T_c$ is the minimum of Eq.~\ref{eq:feature_similarity} over endpoint displacement, path length, and normalized chord deviation. For rotation, compare start-relative rotation paths at 20 samples. The cost of matching two responses $a,b$ is
\begin{equation}
\label{eq:rotation_matching_cost}
C(a,b)
=
\min\left(
1,
\frac{
20^{-1}\sum_{k=1}^{20}
\theta((R_a^{(k)})^\top R_b^{(k)})
+
|\ell_a-\ell_b|
}{
\max((\ell_a+\ell_b)/2,10^\circ)
}
\right),
\end{equation}
where $\theta$ and accumulated angular travels $\ell_a,\ell_b$ are in degrees. Equal-size empirical expansions of the two response sets are optimally assigned, giving $U_c=100(1-\operatorname{mean} C)$ over the assigned pairs. This retains rotation direction and intermediate motion, including overshoot and return.

For channel $h$, let $\rho_{r,c,h}$ be the fraction of role $r$'s responses exceeding its duration-scaled static-noise threshold. With $V_{c,\mathrm{translation}}=T_c$ and $V_{c,\mathrm{rotation}}=U_c$, fairness is
\begin{equation}
\label{eq:coordination_fairness}
S_{\mathrm{fair}}
=
\frac{1}{|\mathcal C|}
\sum_{c\in\mathcal C}
\min_{h\in\mathcal H_c}
\left[
V_{c,h}
\min(\rho_{A,c,h},\rho_{B,c,h})
\right].
\end{equation}

Here $\mathcal C$ contains the action classes with sufficient response evidence for both roles. A case without such evidence is unavailable. Equal nonresponse receives zero, and inactive channels do not inflate the score. The case score averages its eligible action classes; the reported aggregate averages evaluable cases and reports their coverage. This measures comparable commanded responses across roles, not success on a joint cooperative goal.

\paragraph{Cross-view motion consistency.}
A VLM directly compares before-and-after screenshots from the observer and actor views in the 20-case lateral-translation suite. Frames 10 and 60 are used to judge whether the other player's horizontal image position moves left, right, or remains stationary, without requesting localization boxes, coordinates, or visual markers. Each view is checked independently against the expected image-space direction of the recorded action. For binary correctness judgments $q_{\mathrm{observer}},q_{\mathrm{actor}}\in\{0,1\}$,
\begin{equation}
\label{eq:cross_view_motion}
S_{\mathrm{cross\mbox{-}view}}
=
100\,q_{\mathrm{observer}}q_{\mathrm{actor}}.
\end{equation}

Both views must be correct for a case to pass. An unclear judgment or an unparseable judge response is marked unavailable and excluded from the score average. A missing generated video is scored as a task failure with $S_{\mathrm{cross\mbox{-}view}}=0$ and remains in the aggregate. The aggregate averages all scored cases, while judge-unavailable cases are excluded and coverage is reported separately. This direct-VLM protocol remains an experimental visual assessment.

\subsection{Shared world}

\paragraph{Shared-state synchronization.}
A VLM answers independent binary questions: did the intended action succeed; do both views show the same expected number of changed blocks; and did the change occur at the marked target location? Placement adds a fourth question: do both views show the same intended material? Initial and synchronized post-action images supply context. Unclear visual evidence receives zero under this judge protocol. Before voting, a per-frame visibility check requires the marked target cross to be observable; a failed check sets all scored answers for that frame to zero. Each gated question is majority-voted over frames 60, 70, and 80 in the 120-case experiment.
\begin{equation}
\begin{aligned}
S_{\mathrm{place}}
&=
\frac{
q_{\mathrm{success}}
+
q_{\mathrm{count}}
+
q_{\mathrm{location}}
+
q_{\mathrm{material}}
}{4},\\
S_{\mathrm{destroy}}
&=
\frac{
q_{\mathrm{success}}
+
q_{\mathrm{count}}
+
q_{\mathrm{location}}
}{3}.
\end{aligned}
\end{equation}
The overall score is the equal mean of the placement and destruction means. Success is a question, not a multiplicative gate in this protocol.

\paragraph{Shared-state persistence.}
The judge receives labeled before-and-after pairs for the operator and the returning observer. It asks whether the expected change occurred in the operator view and whether the returning observer sees the same expected change at the marked cross. For each pair, the marked cross and its center target cell must be clearly visible in both the before and after images; otherwise the corresponding question is zero. A case uses the strict product of the two binary answers. Placement and destruction pass rates are macro-averaged, so each operation contributes equally.

\paragraph{Cross-view structural consistency.}
Two binary judgments ask whether the generated revealed building is the same building as its corresponding GT, and whether generated A and B show the same building. Each question uses majority vote over frames 80, 100, and 120; the case score is the minimum of the two votes. In the current merged design, occluder removal and target revelation are shared prerequisites.

\paragraph{Spatial reasoning.}
After the same reveal prerequisites, spatial evaluation reuses the GT-identity judgment from structural consistency. It does not ask a second identity question. Segmentation measures projected-area agreement
\[
S_{\mathrm{area}}
=
100
\frac{
\min(A_{\mathrm{pred}},A_{\mathrm{GT}})
}{
\max(A_{\mathrm{pred}},A_{\mathrm{GT}})
}.
\]
Camera-angle agreement decreases linearly with wrapped angular error and reaches zero at $30^\circ$. After passing the prerequisites and identity check, the score is the lower of area and angle agreement. Missing reliable geometry makes the case unavailable; a failed prerequisite is an observed failure and scores zero. The shared prerequisites prevent duplicated identity judgments across structural and spatial evaluation.

\paragraph{Race-condition consistency.}
For simultaneous single-event placement, the VLM reports target visibility, changed-block count, and material separately for A and B at synchronized frames 60, 80, and 100, using the intended candidate blocks as reference images. At checkpoint $t$, let $v_t$ indicate that the target is visible in both views, $a_t$ indicate agreement on count and material, and $u_t$ indicate a valid placement outcome. Valid outcomes contain one or two blocks made from the supplied candidate materials; two-block material combinations are unordered. The case score is
\begin{equation}
S_{\mathrm{race}}
=
100\mathbf{1}
\left[
\sum_{t\in\{60,80,100\}}
v_ta_tu_t
\geq 2
\right].
\end{equation}
Conditions must pass jointly at the same checkpoint before temporal voting. Both views showing an empty target can satisfy state agreement but cannot pass valid placement; a hidden target, an invalid material, or three or more changed blocks fails the checkpoint. Missing judge output is a task error. This evaluates visible shared-state consistency and execution under conflicting actions without prescribing a winner.

\subsection{Revisit consistency}
\paragraph{Style.}
We crop each image to vertical fractions $[.08,.88]$ and horizontal fractions $[.08,.92]$, suppress near-black/white achromatic pixels, and normalize a joint $16\times16\times16$ RGB histogram $h$. For a visit--return pair,
\begin{equation}
\begin{gathered}
d(h_0,h_1)=\sqrt{\tfrac12\mathrm{KL}_2(h_0\Vert m)+
\tfrac12\mathrm{KL}_2(h_1\Vert m)},\\
m=(h_0+h_1)/2.
\end{gathered}
\end{equation}
The generated distance is compared with the engine GT distance for the same pair. With case-configured pairs $\mathcal P$ (one cross-player pair per directional case in the current suite),
\begin{equation}
S_{\mathrm{style}}=100\max\left(0,1-\frac{|\mathcal P|^{-1}
\sum_{p\in\mathcal P}|d^{\mathrm{gen}}_p-d^{\mathrm{GT}}_p|}{0.25}\right).
\label{eq:style}
\end{equation}
For the Engine-GT baseline, $d^{\mathrm{gen}}_p$ is obtained from an
independent native recapture of the same case, while
$d^{\mathrm{GT}}_p$ comes from the original reference recording.
Its score therefore measures agreement in visit--return color change
between two engine recordings and need not equal 100.
This detects changes such as sand-like colors becoming grass-like colors. It is insensitive to spatial rearrangement that preserves the histogram, and is therefore complemented by identity.

\paragraph{Identity.}
For each directional case and evaluated system, the VLM determines
whether the building observed during the initial visit remains the same
building after the delayed return. Engine GT is evaluated using the
same identity question and serves as the reference system, while each
generated system---Solaris (MP), Gamma-World, and Solaris (SP)---is
judged independently under the same protocol. No system-specific
eligibility rule or GT-based score gate is applied.

A case passes only if every generated evidence pair preserves the
building's recognizable identity, including its overall layout,
silhouette, openings, roof, and other distinctive structural
components. The current directional cases contain one evidence pair
per system. Missing or insufficient evidence is treated as a failed
identity judgment rather than replaced by a passing vote. We report
the mean score across the 20 directional cases as the primary
aggregate. As a stricter role-swap diagnostic, we additionally compute
the mean of the lower primary and counter-direction scores within each
of the ten paired scenes.

\subsection{Human Pairwise Evaluation}
\label{sec:human-evaluation}

\paragraph{Data preparation.}
We sample cases from each evaluated capability and collect the
corresponding videos from four systems, denoted by
\begin{equation}
\mathcal{M}
=
\{\text{Engine GT},\text{Solaris (MP)},
\text{Gamma-World},\text{Solaris (SP)}\}.
\end{equation}
For each case and evaluation dimension, we construct a complete
pairwise comparison among the four systems, yielding
\begin{equation}
|\mathcal{P}|
=
\binom{|\mathcal{M}|}{2}
=
\binom{4}{2}
=
6
\end{equation}
system pairs. Using $E$, $S_{\mathrm{MP}}$, $G$, and $S_{\mathrm{SP}}$
to denote Engine GT, Solaris (MP), Gamma-World, and Solaris (SP),
respectively, the comparison set is
\begin{equation}
\mathcal{P}
=
\{
(E,S_{\mathrm{MP}}),
(E,G),
(E,S_{\mathrm{SP}}),
(S_{\mathrm{MP}},G),
(S_{\mathrm{MP}},S_{\mathrm{SP}}),
(G,S_{\mathrm{SP}})
\}.
\end{equation}
Each annotation task compares one pair of outputs generated under the
same benchmark case and task condition, ensuring that both videos
correspond to the same evaluation target.

\paragraph{Annotation protocol.}
Each task presents two videos with synchronized playback controls and
a dimension-specific evaluation question; system identities are
hidden, while video positions and the order of pairwise comparison
tasks are randomized. The instructions specify the evidence to
consider and unrelated factors to ignore. For example, when evaluating
control isolation, annotators determine whether the target player's
response remains consistent when its own action is fixed and the other
player's action changes, including whether an idle target view remains
stable. Differences in texture sharpness, general visual quality, and
minor rendering details are ignored unless they directly affect the
target capability.

After viewing both videos, the annotator selects one of four outcomes:
\begin{equation}
y_{c,d}^{(i,j)}
\in
\{
i \succ j,\;
i \sim j,\;
j \succ i,\;
\varnothing
\},
\qquad
(i,j)\in\mathcal{P},
\end{equation}
where $c$ denotes the benchmark case, $d$ denotes the evaluation
dimension, $i\succ j$ and $j\succ i$ indicate that one system
outperforms the other, and $i\sim j$ indicates no meaningful
difference. The symbol $\varnothing$ denotes an invalid comparison for
which a reliable judgment cannot be made because of missing content,
playback failure, or insufficient visual evidence. Because each task
evaluates only one capability dimension, the same video pair may
receive different judgments under different evaluation criteria.

\newpage
\section{Experiments}
\label{sec:experiments}

\subsection{Setup and evidence}

We evaluate Solaris (MP), Gamma-World, and Solaris (SP) across three
benchmark tracks, using Engine GT as the reference. The evaluation
comprises 495 case configurations covering ten capabilities, with all
scores reported on a 0--100 scale. Track averages weight their
constituent metrics equally, and the overall score is the equally
weighted average of all ten capability scores.

\begin{figure}[H]
    \centering

    \begin{subfigure}[t]{0.57\linewidth}
        \centering
        \includegraphics[width=\linewidth]
        {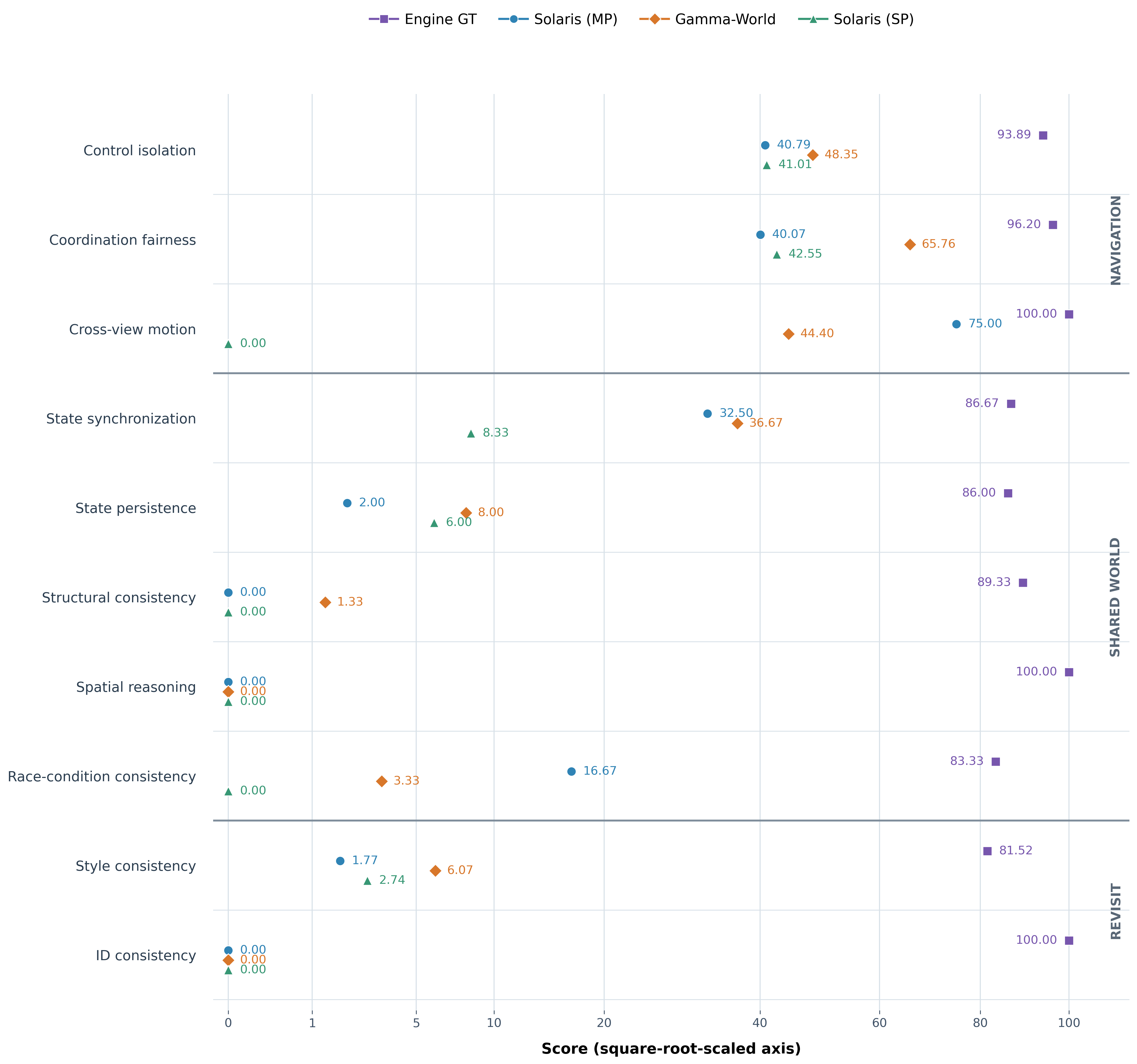}
        \caption{Ten-capability score profile.}
        \label{fig:capability-profile}
    \end{subfigure}
    \hfill
    \begin{subfigure}[t]{0.41\linewidth}
        \centering
        \includegraphics[width=\linewidth]
        {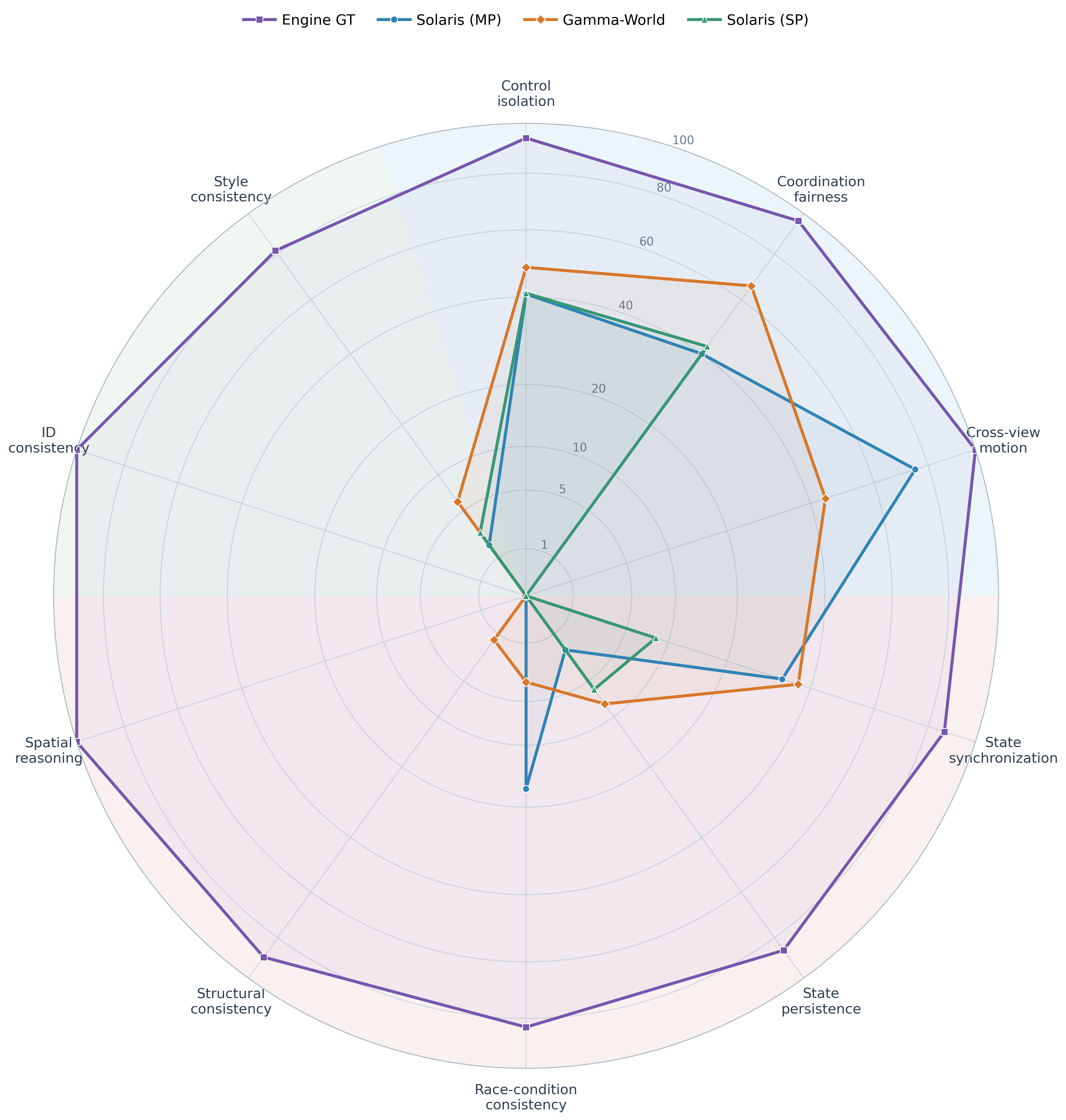}
        \caption{Ten-capability radar profile.}
        \label{fig:capability-radar}
    \end{subfigure}

    \caption{
        Capability comparison across Navigation, Shared World, and
        Revisit for Engine GT, Solaris (MP), Gamma-World, and
        Solaris (SP). Both panels use square-root-scaled positions to
        improve the visibility of differences among low scores;
        displayed tick labels and annotations retain the original
        0--100 scores.
    }
    \label{fig:capability-overview}
\end{figure}

Table~\ref{tab:scorecard} and Figure~\ref{fig:capability-overview}
summarize the results. Among the generated systems, Gamma-World
achieves the highest overall average (21.39), followed closely by
Solaris (MP) (20.88), while Solaris (SP) obtains 10.06.

On Navigation, Gamma-World narrowly leads Solaris (MP) in track
average (52.84 vs.\ 51.95). Gamma-World performs better on control
isolation (48.35 vs.\ 40.79) and coordination fairness
(65.76 vs.\ 40.07), whereas Solaris (MP) has a clear advantage in
cross-view motion consistency (75.00 vs.\ 44.40).

Solaris (MP) slightly leads Gamma-World on Shared World
(10.23 vs.\ 9.87), primarily because of its stronger race-condition
consistency (16.67 vs.\ 3.33). Solaris (SP) obtains a Shared World
average of 2.87 and scores 0.00 on race-condition consistency.

All three generated systems perform poorly on Revisit, with track
averages of 3.04 for Gamma-World, 1.37 for Solaris (SP), and 0.89 for
Solaris (MP). These near-zero scores indicate that none of the
evaluated systems exhibits reliable revisit consistency, particularly
in preserving visual style and building identity after a delayed
cross-player return.

\begin{table}[H]
\centering
\caption{
    \textbf{Main results across ten capabilities.}
    Scores are displayed on a 0--100 scale; higher is better.
    Track averages weight their constituent metrics equally, and the
    overall score weights all ten capabilities equally. Bold marks the
    highest generated-system score or a tie within each capability;
    averages are bold for readability.
}
\label{tab:scorecard}
\begin{tabular}{@{}lrrrrr@{}}
\toprule
Capability & Cases & Engine GT & Solaris (MP) & Gamma-World
& Solaris (SP) \\
\midrule
\multicolumn{6}{@{}l}{\textit{Navigation}} \\
Control isolation
    & 180 & 93.89 & 40.79 & \textbf{48.35} & 41.01 \\
Coordination fairness
    & 180 & 96.20 & 40.07 & \textbf{65.76} & 42.55 \\
Cross-view motion
    & 20 & 100.00 & \textbf{75.00} & 44.40 & 0.00 \\
\textbf{Track average}
    & -- & \textbf{96.70} & \textbf{51.95} & \textbf{52.84}
    & \textbf{27.85} \\
\midrule
\multicolumn{6}{@{}l}{\textit{Shared world}} \\
State synchronization
    & 120 & 86.67 & 32.50 & \textbf{36.67} & 8.33 \\
State persistence
    & 50 & 86.00 & 2.00 & \textbf{8.00} & 6.00 \\
Structural consistency
    & 75 & 89.33 & 0.00 & \textbf{1.33} & 0.00 \\
Spatial reasoning
    & 75 & 100.00 & \textbf{0.00} & \textbf{0.00}
    & \textbf{0.00} \\
Race-condition consistency
    & 30 & 83.33 & \textbf{16.67} & 3.33 & 0.00 \\
\textbf{Track average}
    & -- & \textbf{89.07} & \textbf{10.23} & \textbf{9.87}
    & \textbf{2.87} \\
\midrule
\multicolumn{6}{@{}l}{\textit{Revisit}} \\
Style consistency
    & 20 & 81.52 & 1.77 & \textbf{6.07} & 2.74 \\
ID consistency
    & 20 & 100.00 & \textbf{0.00} & \textbf{0.00}
    & \textbf{0.00} \\
\textbf{Track average}
    & -- & \textbf{90.76} & \textbf{0.89} & \textbf{3.04}
    & \textbf{1.37} \\
\midrule
\textbf{Overall average}
    & -- & \textbf{91.69} & \textbf{20.88} & \textbf{21.39}
    & \textbf{10.06} \\
\bottomrule
\end{tabular}
\end{table}

\subsection{Per-Dimension Results}
\label{sec:per-dimension-results}

We discuss the principal results across navigation, shared-world
consistency, and delayed revisit below. 
Representative qualitative examples are provided in
Appendix~\ref{app:qualitative-examples}.

\subsubsection{Navigation}

\paragraph{Control isolation.}
Among the generated systems, Gamma-World (48.35) leads
Solaris (SP) (41.01) and Solaris (MP) (40.79), but remains below
Engine GT (93.89). Figure~\ref{fig:app-control-isolation}
illustrates the central failure: although Bravo receives no command,
Alpha's action still affects Bravo's view. Gamma-World exhibits a
smaller unintended shift in this example, while both Solaris models
change the inactive view more substantially, producing larger and
less consistent viewpoint and scene changes even in the absence of
a command to Bravo.

\paragraph{Coordination fairness.}
Gamma-World (65.76) outperforms Solaris (SP) (42.55) and
Solaris (MP) (40.07), but remains below Engine GT (96.20).
Among the generated systems, Gamma-World gives Alpha and Bravo the
most comparable camera-motion responses to the same command. Both
Solaris models show larger role-dependent differences in motion
magnitude, resulting in less balanced commanded responses across
the two player views.

\paragraph{Cross-view motion consistency.}
Across the 20 cross-view motion cases, Solaris (MP) (75.00) leads
Gamma-World (44.40) and Solaris (SP) (0.00), while Engine GT
(100.00) provides the reference. Appendix
Figure~\ref{fig:cross-view-motion-app} illustrates a representative
result: Engine GT and Solaris (MP) reproduce the expected displacement
in both the actor and observer views, whereas Gamma-World and
Solaris (SP) show little corresponding change and therefore receive
lower scores.

\subsubsection{Shared World}

\paragraph{State synchronization.}
Engine GT scores 86.67, compared with 36.67 for Gamma-World, 32.50
for Solaris (MP), and 8.33 for Solaris (SP).
Figure~\ref{fig:app-state-sync} shows a representative destruction
case. The target block should be removed at the same location in both
the Alpha and Bravo views. Engine GT maintains the most consistent
shared result. The generated systems more frequently produce
cross-view disagreement, incomplete destruction, or a changed target
state, with Solaris (SP) obtaining the lowest synchronization score.

\paragraph{State persistence.}
Engine GT scores
86.00, while Gamma-World, Solaris (SP), and Solaris (MP) score 8.00,
6.00, and 2.00, respectively. As illustrated in
Figure~\ref{fig:app-state-persistence}, Gamma-World at least preserves
the cross-shaped ground marker after the return, although it does not
reliably preserve the intended destroyed state. In contrast, both
Solaris models lose the cross-shaped marker and return to scenes that
no longer correspond to the edited region. Gamma-World therefore
achieves the highest generated-system score, but all three generated
systems remain far below Engine GT.

\paragraph{Structural consistency.}
Engine GT obtains 89.33, while Gamma-World scores 1.33 and both
Solaris (MP) and Solaris (SP) score 0.00.
In Figure~\ref{fig:app-structural-spatial}, removing the occluder
should reveal the same building that is already visible to Bravo.
Engine GT preserves the shared structure across the two views.
The generated outputs instead leave Alpha facing altered geometry,
an obstructing surface, or a scene that does not correspond to
Bravo's view.

\paragraph{Spatial reasoning.}
Engine GT reaches 100.00, whereas Solaris (MP), Gamma-World, and
Solaris (SP) all score 0.00.
The building-reveal example in
Figure~\ref{fig:app-structural-spatial} illustrates why structural
agreement is necessary for spatial reasoning. If the target building
is missing, remains occluded, or changes across views, its relative
direction cannot be recovered consistently from the two-player
observations.

\paragraph{Race-condition consistency.}
Solaris (MP) achieves 16.67, followed by Gamma-World at 3.33 and
Solaris (SP) at 0.00, while Engine GT reaches 83.33.
The task asks Alpha and Bravo to place different materials
concurrently at the same target location, after which both views
should show one consistent final state.
Figure~\ref{fig:app-race-condition} shows that Solaris (MP) produces
more similar results across the two views than the other generated
systems, although the resulting material can still differ from the
requested one. Gamma-World exhibits greater disagreement in the
resulting block state. Solaris (SP) independently places a block of
the player-specific color in each view, producing two different
outcomes and therefore showing no cross-view consistency.

\subsubsection{Revisit}

\paragraph{Style consistency.}
Gamma-World obtains the highest generated-system score of 6.07,
followed by Solaris (SP) at 2.74 and Solaris (MP) at 1.77.
Engine GT reaches 81.52.
Figure~\ref{fig:app-revisit-style-id} compares the initial Alpha view
with Bravo's delayed return. Gamma-World retains more of the coarse
visual appearance than the two Solaris variants in this example, but
all generated-system scores remain low. Similar colors or textures
alone are therefore insufficient to demonstrate that the same world
has been preserved.

\paragraph{ID consistency.}
Engine GT scores 100.00, while Solaris (MP), Gamma-World, and
Solaris (SP) all score 0.00.
In Figure~\ref{fig:app-revisit-style-id}, the building visible in the
initial view should remain recognizable after the delayed return.
The generated systems remove, alter, or replace the relevant structure,
preventing the returning view from preserving the building's identity.
Thus, partial style similarity does not imply identity consistency.

\subsection{Human Preference Results and Metric Alignment}
\label{sec:human-preference-results}

\paragraph{Human preference win ratio.}
Following VBench~\cite{huang2024vbench}, a win, tie, and loss contribute
$1$, $0.5$, and $0$ points, respectively. Let $\mathcal{J}_{m,d}$
denote the set of valid pairwise judgments involving system $m$ in
evaluation dimension $d$. The human preference win ratio is defined as
\begin{equation}
H_{m,d}
=
\frac{1}{|\mathcal{J}_{m,d}|}
\sum_{j\in\mathcal{J}_{m,d}} h_{m,d}^{(j)},
\qquad
h_{m,d}^{(j)} \in \{0,0.5,1\}.
\end{equation}
We summarize performance across the evaluated dimensions using the
macro-average
\begin{equation}
\overline{H}_{m}
=
\frac{1}{|\mathcal{D}|}
\sum_{d\in\mathcal{D}} H_{m,d},
\end{equation}
where $\mathcal{D}$ denotes the set of ten evaluated dimensions.
Engine GT ($98.67\%$) achieves the highest overall human win ratio.
Among the generated systems, Gamma-World ($46.00\%$) ranks first,
followed by Solaris (MP) ($32.00\%$) and Solaris (SP) ($23.33\%$).
Gamma-World achieves the highest or joint-highest human win ratio among
the generated systems in eight of the ten dimensions, whereas Solaris
(MP) leads in cross-view motion and race-condition consistency.
This result is consistent with the overall automatic evaluation, which
also ranks Gamma-World above Solaris (MP) and Solaris (SP).

\noindent
\begin{minipage}{\linewidth}
\centering
\includegraphics[
    width=\linewidth,
    keepaspectratio
]{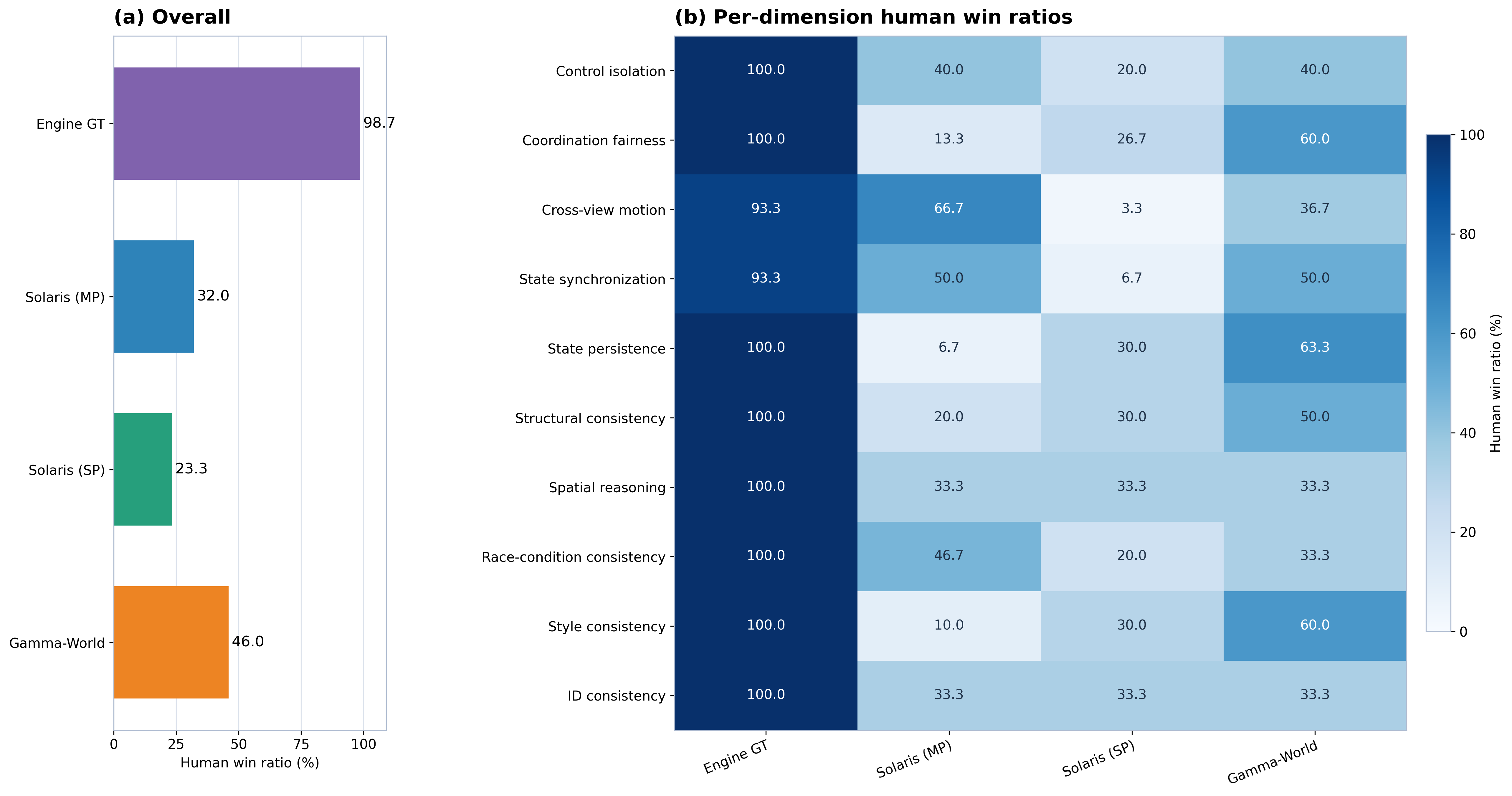}
\captionof{figure}{
\textbf{Human preference win ratios.}
The left panel reports the macro-average win ratio across the ten
evaluated dimensions. The right panel reports the win ratio of each
system within each dimension.
}
\label{fig:human-win-ratios}
\end{minipage}

\paragraph{Alignment with automatic evaluation.}
Following VBench~\cite{huang2024vbench}, we convert the automatic
scores into system-level win ratios. Let $\mathcal{M}$ denote the set
of four evaluated systems, including Engine GT, and let $q_{m,d}$
denote the automatic score of system $m$ in dimension $d$.
For every pair of distinct systems $m$ and $m'$, we define
\begin{equation}
a_{m,m',d}
=
\begin{cases}
1,   & q_{m,d} > q_{m',d},\\
0.5, & q_{m,d} = q_{m',d},\\
0,   & q_{m,d} < q_{m',d},
\end{cases}
\end{equation}
and calculate the automatic-evaluation win ratio as
\begin{equation}
A_{m,d}
=
\frac{1}{|\mathcal{M}|-1}
\sum_{m'\in\mathcal{M}\setminus\{m\}}
a_{m,m',d}.
\end{equation}
For each dimension, we then calculate Spearman's rank correlation
between the human and automatic win ratios, assigning average ranks
to tied values:
\begin{equation}
\rho_d
=
\operatorname{Spearman}
\left(
\{H_{m,d}\}_{m\in\mathcal{M}},
\{A_{m,d}\}_{m\in\mathcal{M}}
\right).
\end{equation}
As shown in Figure~\ref{fig:human-metric-alignment}, the correlations
are generally high across the ten dimensions (range: $0.63$--$1.00$;
mean: $0.95$).
Seven dimensions exhibit identical system-level rankings between
human preferences and the automatic evaluation: coordination fairness
($\rho=1.00$), cross-view motion ($\rho=1.00$), state persistence
($\rho=1.00$), spatial reasoning ($\rho=1.00$), race-condition
consistency ($\rho=1.00$), style consistency ($\rho=1.00$), and ID
consistency ($\rho=1.00$).
State synchronization ($\rho=0.95$) and structural consistency
($\rho=0.95$) also show strong rank agreement, while control isolation
shows comparatively weaker agreement ($\rho=0.63$). This lower correlation is mainly attributable to near-tied results among the generated systems, for which small changes in relative ranking have a comparatively large effect when only four systems are evaluated. Overall, the automatic evaluation produces system rankings that are
broadly aligned with the aggregated human preferences across the
evaluated dimensions. 

\noindent
\begin{minipage}{\linewidth}
\centering
\includegraphics[
    width=\linewidth,
    keepaspectratio
]{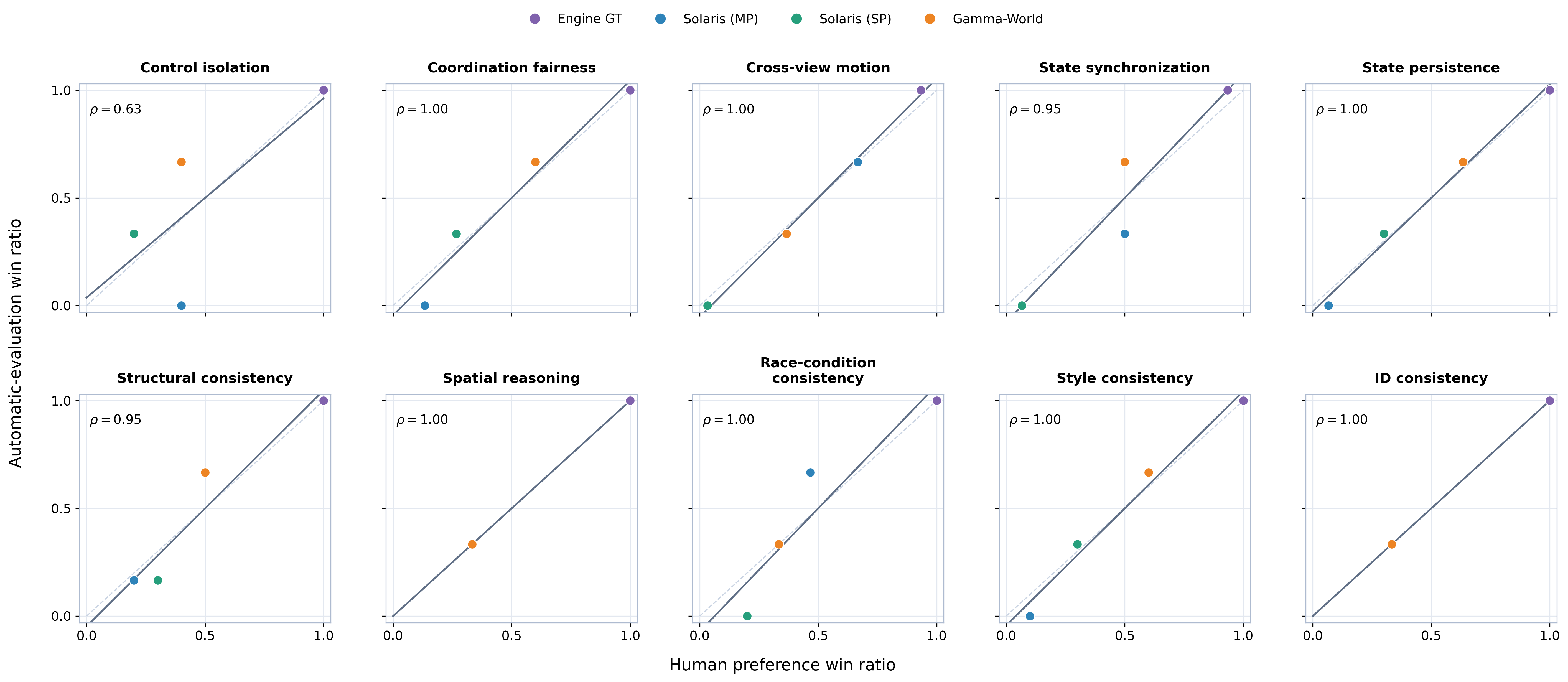}
\captionof{figure}{
\textbf{Alignment between human preference and automatic evaluation.}
Each panel corresponds to one of the ten evaluation dimensions, and
each point represents one system. The horizontal and vertical axes
show the human and automatic-evaluation win ratios, respectively.
The solid line shows a linear fit, and $\rho$ denotes Spearman's
rank correlation.
}
\label{fig:human-metric-alignment}
\end{minipage}

\section{Conclusion and Limitations}
\label{sec:conclusion}

\noindent
\textbf{Multiplayer world models better coordinate information across
independently controlled views, but still fail to maintain a shared world.}
We introduced \bench{} to evaluate whether independently controlled
video streams behave as observations of one shared and persistent
world. Solaris (SP) performs substantially worse than Solaris (MP) on
state synchronization, cross-view motion, and race-condition
consistency, all of which require information about actions or state
changes to be coordinated across views. Solaris (MP) and Gamma-World
generally produce more recognizable player-specific camera motion and
are more likely to propagate an interaction or converge to a visually
consistent outcome across both views. This comparison suggests that
multiplayer-oriented generation provides a better foundation for
coordinating independently controlled views. Nevertheless, both
multiplayer models remain far behind Engine GT and continue to exhibit
substantial cross-view inconsistencies.

\par\medskip
\noindent
\textbf{Failures in basic movement and placement propagate to more
complex tasks.}
Solaris (MP) handles forward and backward translation comparatively
well, but its rotation magnitude and stopping behavior are poorly
calibrated: it may turn insufficiently or continue rotating after the
command has ended. Gamma-World usually follows the intended motion
direction, but its movement magnitude may be too small or its rotation
may overshoot. Both multiplayer systems also struggle with placement.
Solaris (MP) may generate a floating, incomplete, or temporally
desynchronized block, whereas Gamma-World may place the block at the
wrong location, propagate it to only one view, or even convert a
placement instruction into destruction. These failures extend beyond
the corresponding navigation or synchronization dimensions.
Persistence, building reveal, spatial reasoning, race-condition
resolution, and delayed revisit all depend on successful movement and
the creation of a valid initial world state. When these prerequisites
fail, the subsequent task necessarily fails as well. The results
therefore reveal an end-to-end fragility in which errors in basic
control and interaction propagate into more complex shared-world
evaluations.

\par\medskip
\noindent
\textbf{Current systems demonstrate little reliable world memory.}
After an edited region or building leaves view, the generated systems
rarely recover the same object, spatial arrangement, and shared state
when another player returns. Gamma-World sometimes preserves more of
the coarse scene context or color palette than Solaris (MP) and
Solaris (SP), but this relative advantage does not amount to object or
world-state memory: the edited block or target building is still
frequently missing. Moreover, some apparent persistence failures begin
earlier, because the requested edit is never successfully created or
the camera does not turn far enough for the target to leave view. Such
cases cannot be interpreted purely as forgetting, because the
prerequisites for testing memory have not been satisfied. Overall,
current systems may maintain short-term visual continuity, but they do
not yet preserve a stable latent world across actions, viewpoints, and
time.

\paragraph{Limitations.}
Our evaluation is limited to Minecraft. Although Minecraft is widely
used for training and evaluating current world models, its discrete
voxel-based visual style remains substantially different from the
continuous, photorealistic dynamics of the real world. In addition,
future evaluations should include more cases, environments, and action
sequences to test whether the observed results generalize across a
broader range of conditions.

\newpage
{\small
\bibliographystyle{ieeenat_fullname}
\bibliography{multiworldbench}
}
\clearpage
\onecolumn
\appendix

\section{Dataset Diversity Gallery}
\label{app:dataset-gallery}

\begin{center}
    \includegraphics[
        width=\textwidth,
        height=0.9\textheight,
        keepaspectratio
    ]{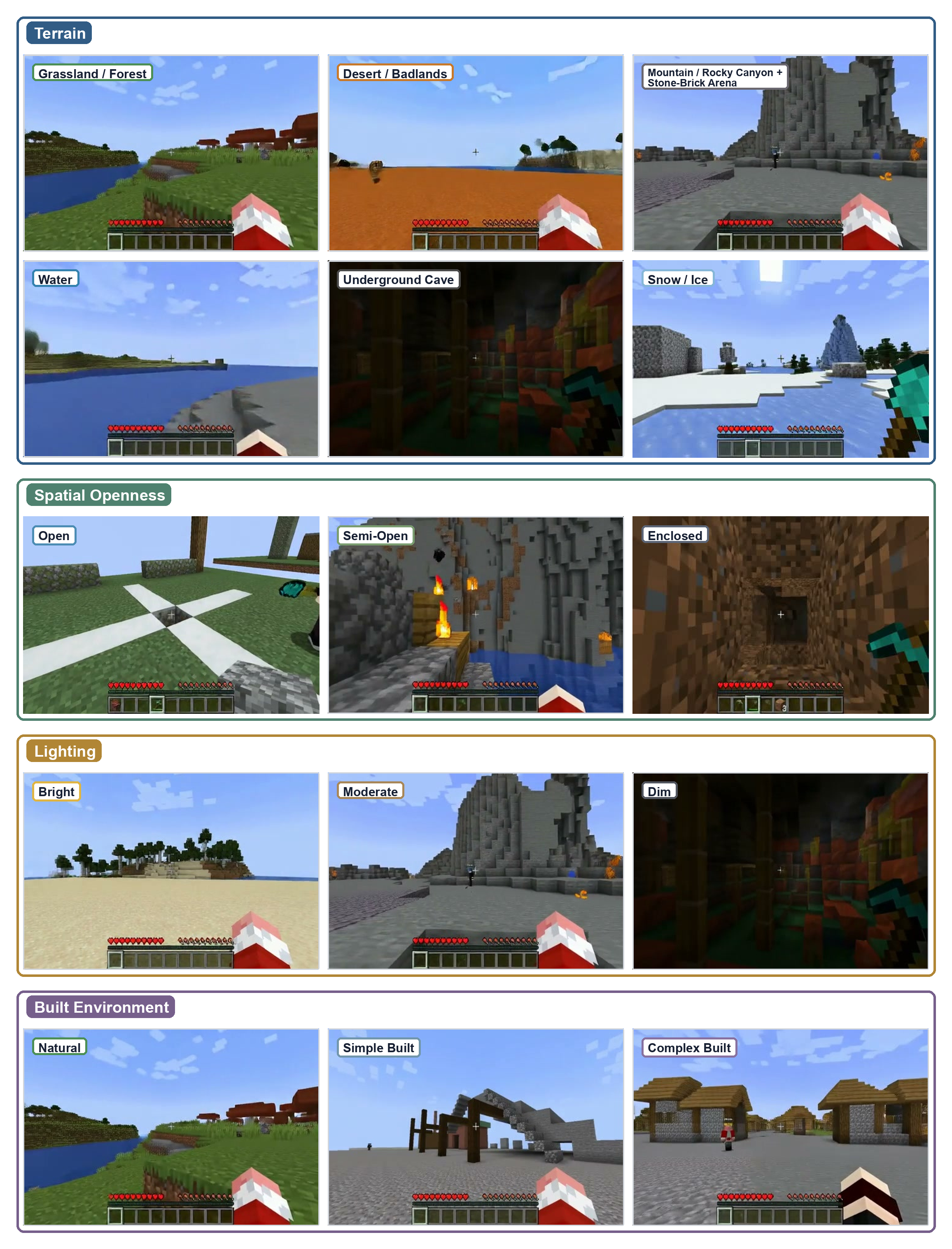}

    \captionof{figure}{
        \textbf{Dataset diversity in \textsc{MultiWorldBench}.}
        Representative environments grouped by terrain, spatial openness,
        lighting condition, and built-environment complexity.
    }
    \label{fig:appendix-dataset-gallery}
\end{center}

\newpage
\clearpage
\onecolumn

\section{Qualitative Examples}
\label{app:qualitative-examples}

This appendix provides representative qualitative examples for the
capabilities discussed in Section~\ref{sec:per-dimension-results}.
The quantitative results are reported in
Table~\ref{tab:scorecard}.

\subsection{Navigation}

\paragraph{Control isolation.}
Figure~\ref{fig:app-control-isolation} compares the inactive Bravo
view before and after Alpha performs the commanded action.

\noindent
\begin{minipage}{\textwidth}
    \centering
    \includegraphics[
        width=\textwidth,
        height=0.8\textheight,
        keepaspectratio
    ]{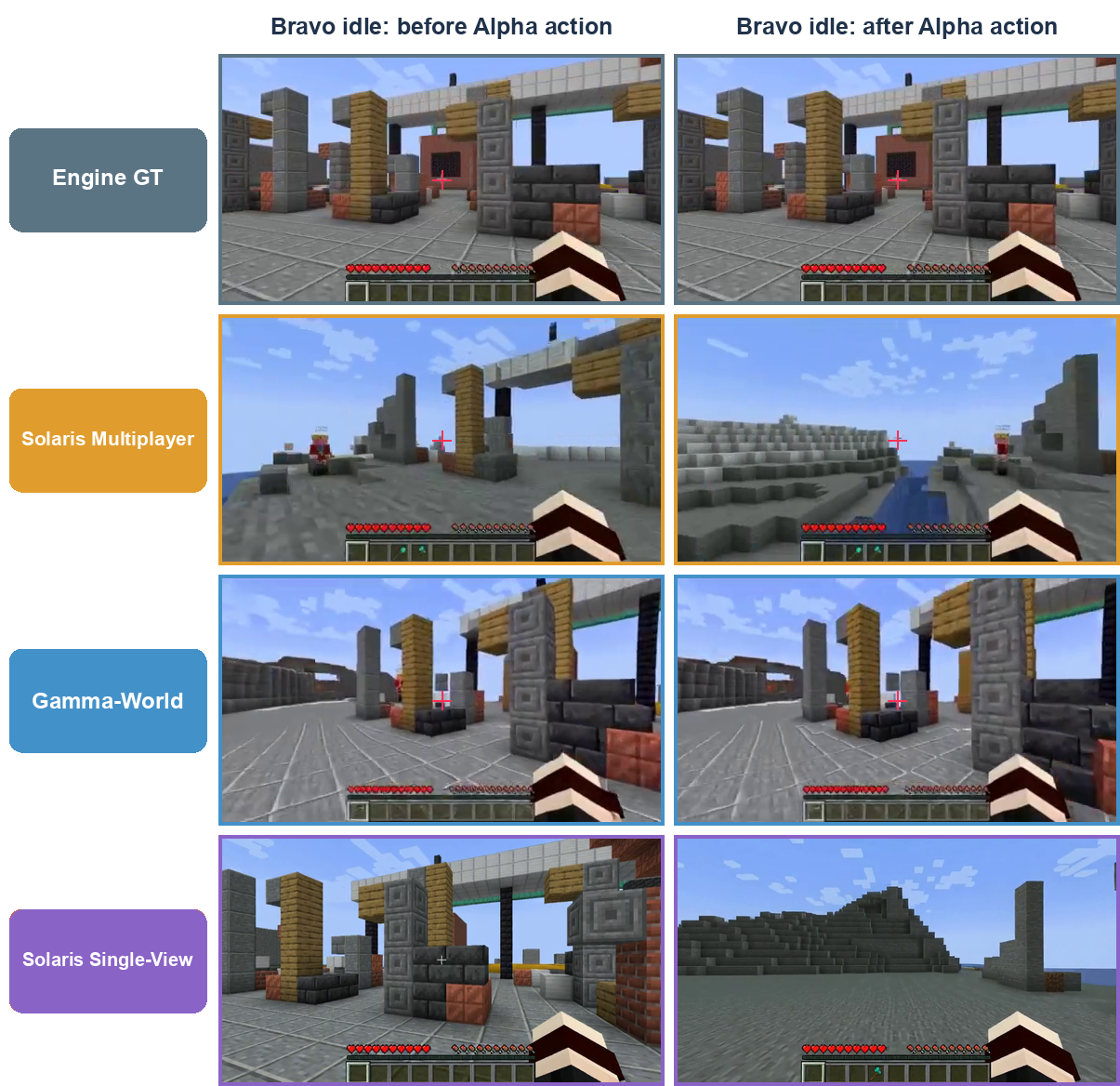}

    \captionof{figure}{
        \textbf{Control isolation}
        (case \texttt{MWB3-N-RG180-NV-SC05-02}).
        The inactive Bravo view is shown before and after Alpha
        performs the commanded action. Rows correspond to Engine GT,
        Solaris (MP), Gamma-World, and Solaris (SP).
    }
    \label{fig:app-control-isolation}
\end{minipage}

\begin{figure}[H]
    \centering
    \includegraphics[width=\linewidth]
    {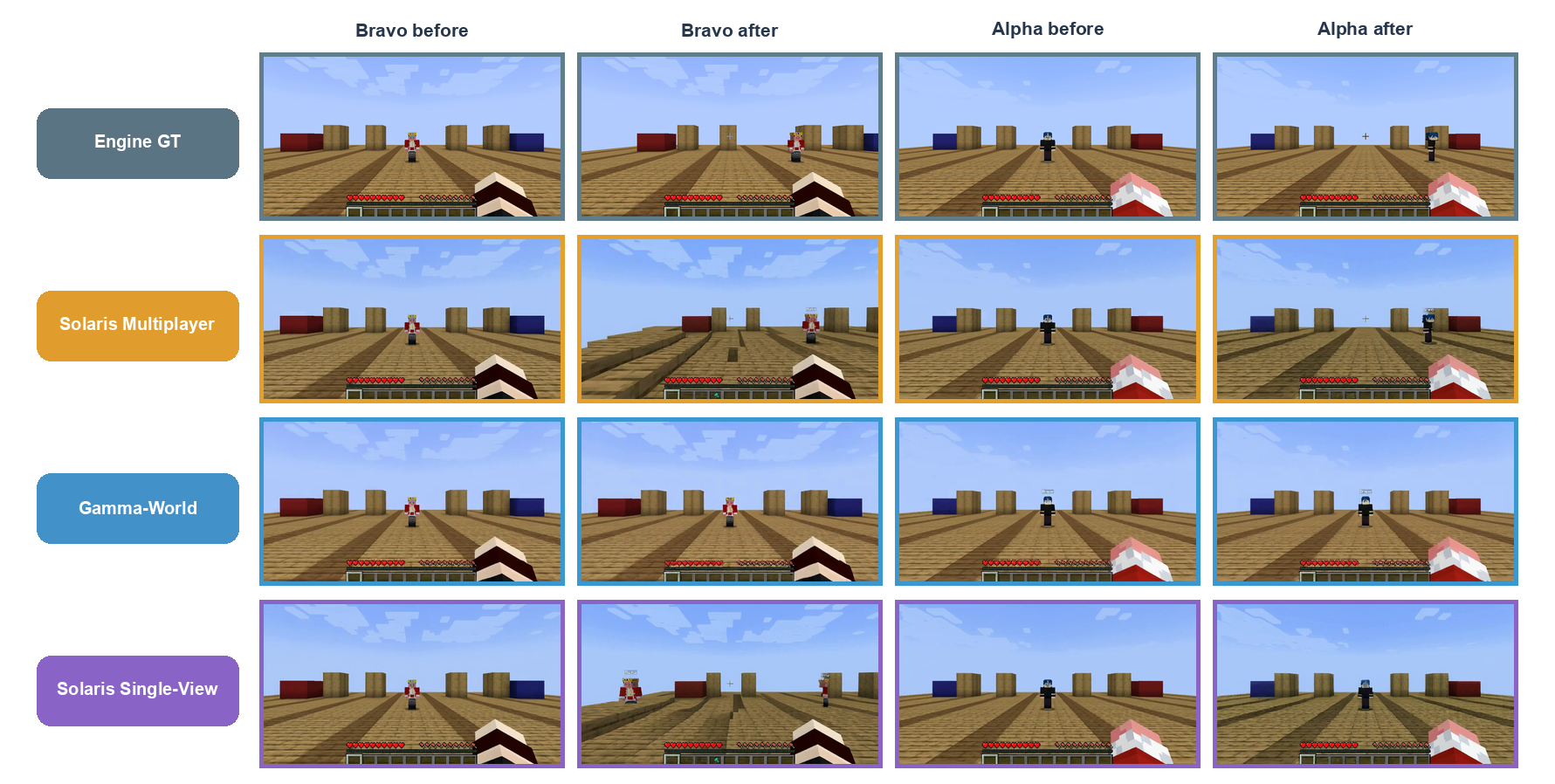}
    \caption{Cross-view motion consistency
    (case MWB4-CV-TR20-S03-03). The actor and observer views are shown before
    and after the action. Rows correspond to Engine GT, Solaris (MP),
    Gamma-World, and Solaris (SP).}
    \label{fig:cross-view-motion-app}
\end{figure}

\paragraph{Cross-view motion.}
Figure~\ref{fig:cross-view-motion-app} compares the actor and observer views
before and after Bravo performs a left strafe while Alpha remains stationary.
The other player should shift right in both views.

\newpage
\subsection{Shared-World Consistency}

\paragraph{State synchronization.}
Figure~\ref{fig:app-state-sync} shows the Alpha and Bravo views before
and after destruction.

\par\medskip

\noindent
\begin{minipage}{\textwidth}
    \centering

    \vspace{3pt}

    \includegraphics[
        width=\textwidth,
        height=0.35\textheight,
        keepaspectratio
    ]{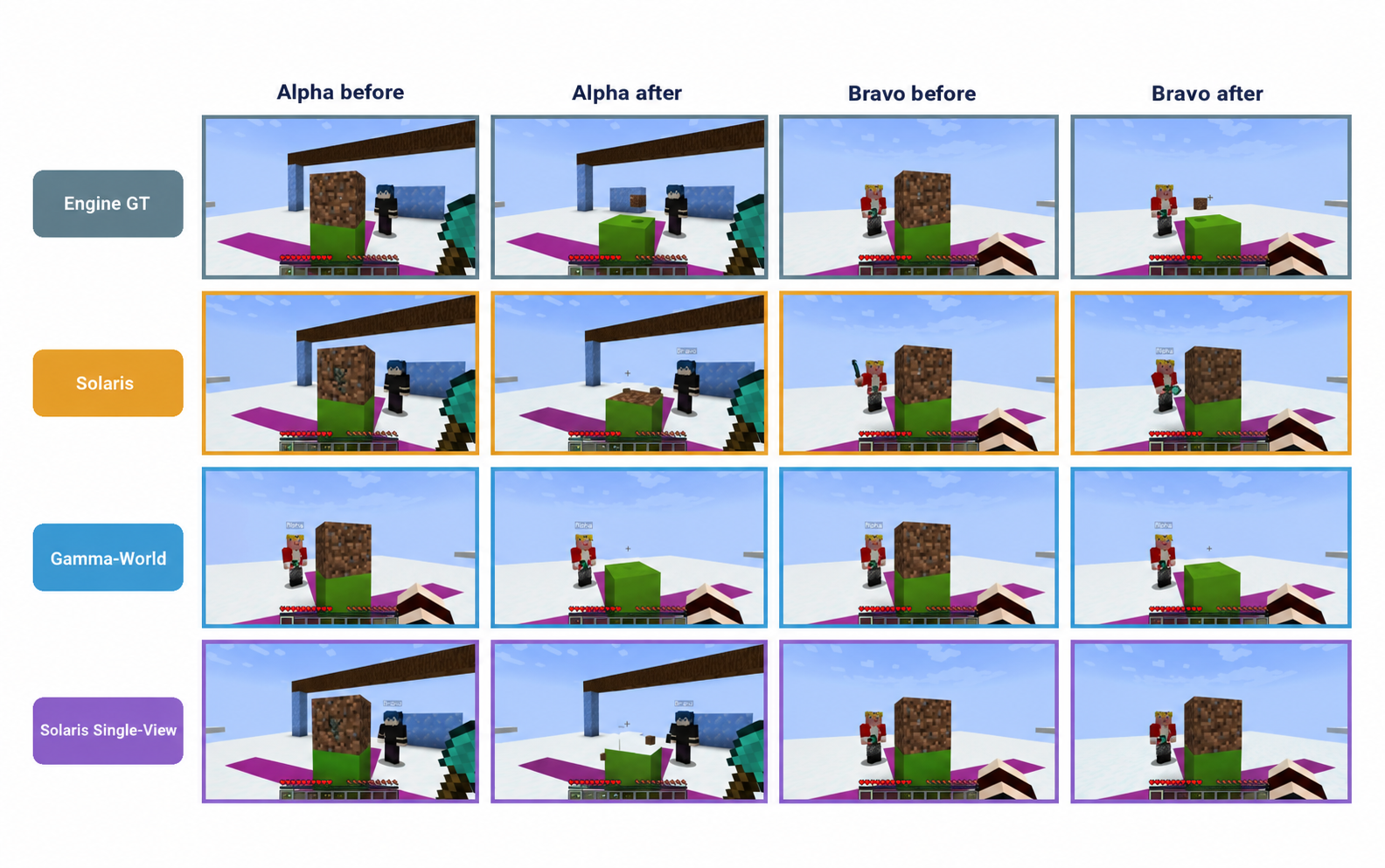}

    \captionof{figure}{
        \textbf{State synchronization under placement and destruction.}
        Destruction case \texttt{MWB3-S-SS003},
        showing both views before and after Alpha attempts to destroy
        the target block. In both panels, rows correspond to Engine GT,
        Solaris (MP), Gamma-World, and Solaris (SP).
    }
    \label{fig:app-state-sync}
\end{minipage}

\clearpage
\paragraph{State persistence.}
Figure~\ref{fig:app-state-persistence} compares the operator's view
after the edit with the other player's view after returning to the
location.

\noindent
\begin{minipage}{\textwidth}
    \centering
    \includegraphics[
        width=\textwidth,
        height=0.78\textheight,
        keepaspectratio
    ]{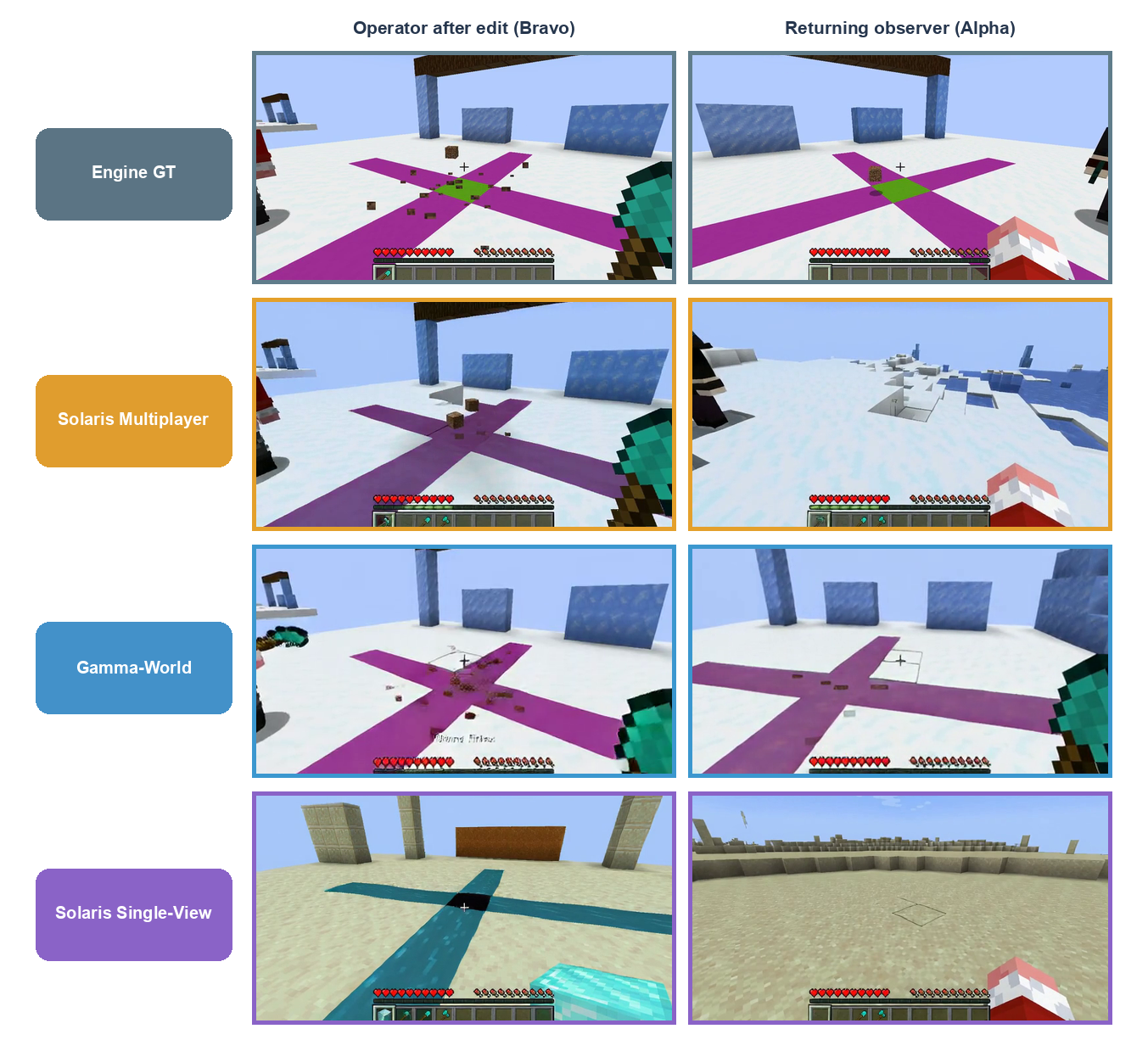}

    \captionof{figure}{
        \textbf{State persistence}
        (case \texttt{MWB3-S-SP001}).
        The left column shows the operator after the edit, and the
        right column shows the other player after returning. Rows
        correspond to Engine GT, Solaris (MP), Gamma-World,
        and Solaris (SP).
    }
    \label{fig:app-state-persistence}
\end{minipage}

\clearpage
\paragraph{Structural consistency and spatial reasoning.}
Figure~\ref{fig:app-structural-spatial} compares whether the building
visible to Bravo is revealed from Alpha after the occluder is removed.

\noindent
\begin{minipage}{\textwidth}
    \centering
    \includegraphics[
        width=\textwidth,
        height=0.78\textheight,
        keepaspectratio
    ]{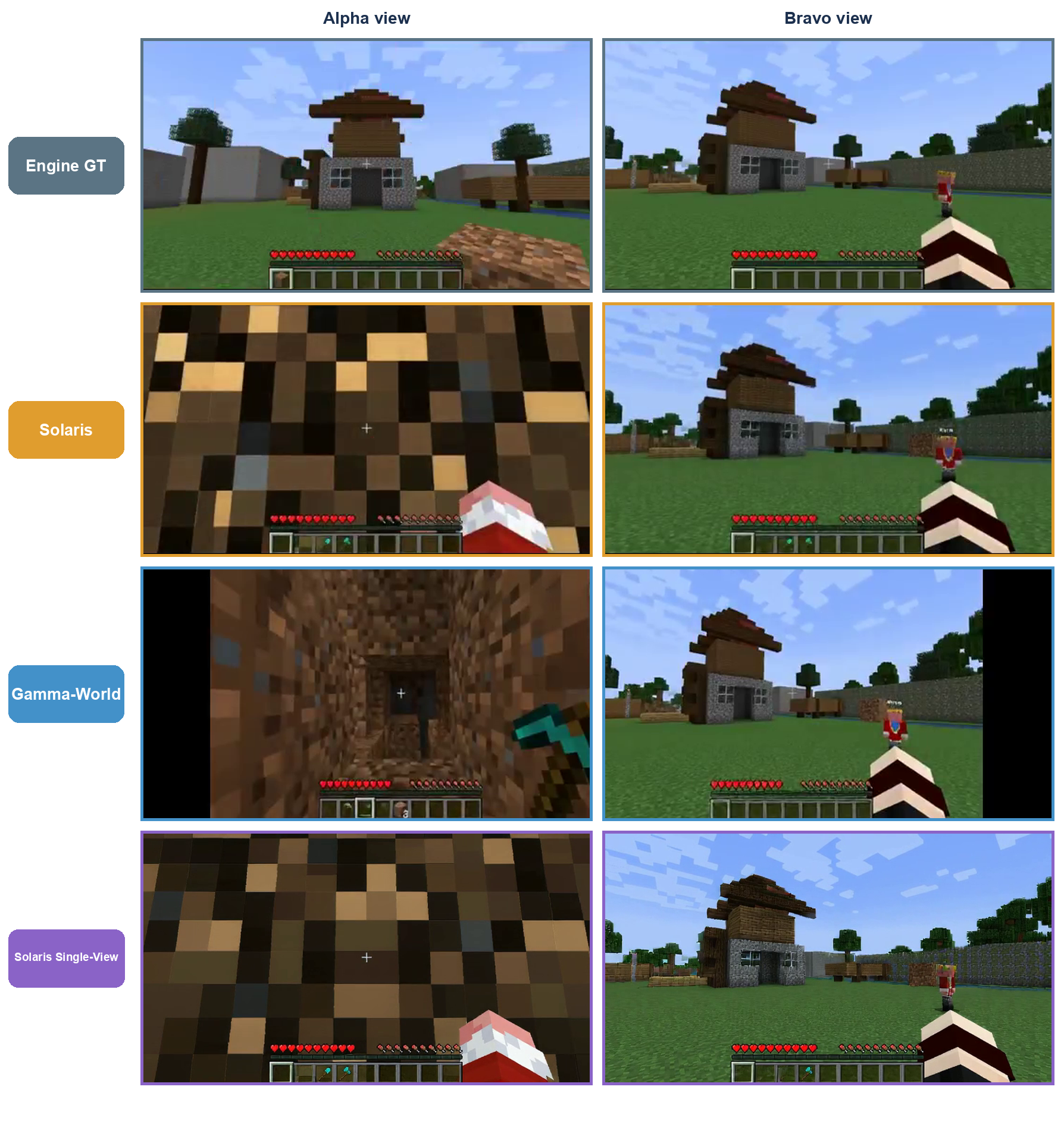}

    \captionof{figure}{
        \textbf{Structural consistency and spatial reasoning}
        (case \texttt{MWB3-S-RV-S01-B01-A20}).
        The Alpha and Bravo views are shown after the
        occluder-removal action. Rows correspond to Engine GT,
        Solaris (MP), Gamma-World, and Solaris (SP).
    }
    \label{fig:app-structural-spatial}
\end{minipage}

\clearpage
\paragraph{Race-condition consistency.}
Figure~\ref{fig:app-race-condition} shows the two player views after
gold and diamond blocks are placed concurrently at the same target
location.

\noindent
\begin{minipage}{\textwidth}
    \centering
    \includegraphics[
        width=\textwidth,
        height=0.78\textheight,
        keepaspectratio
    ]{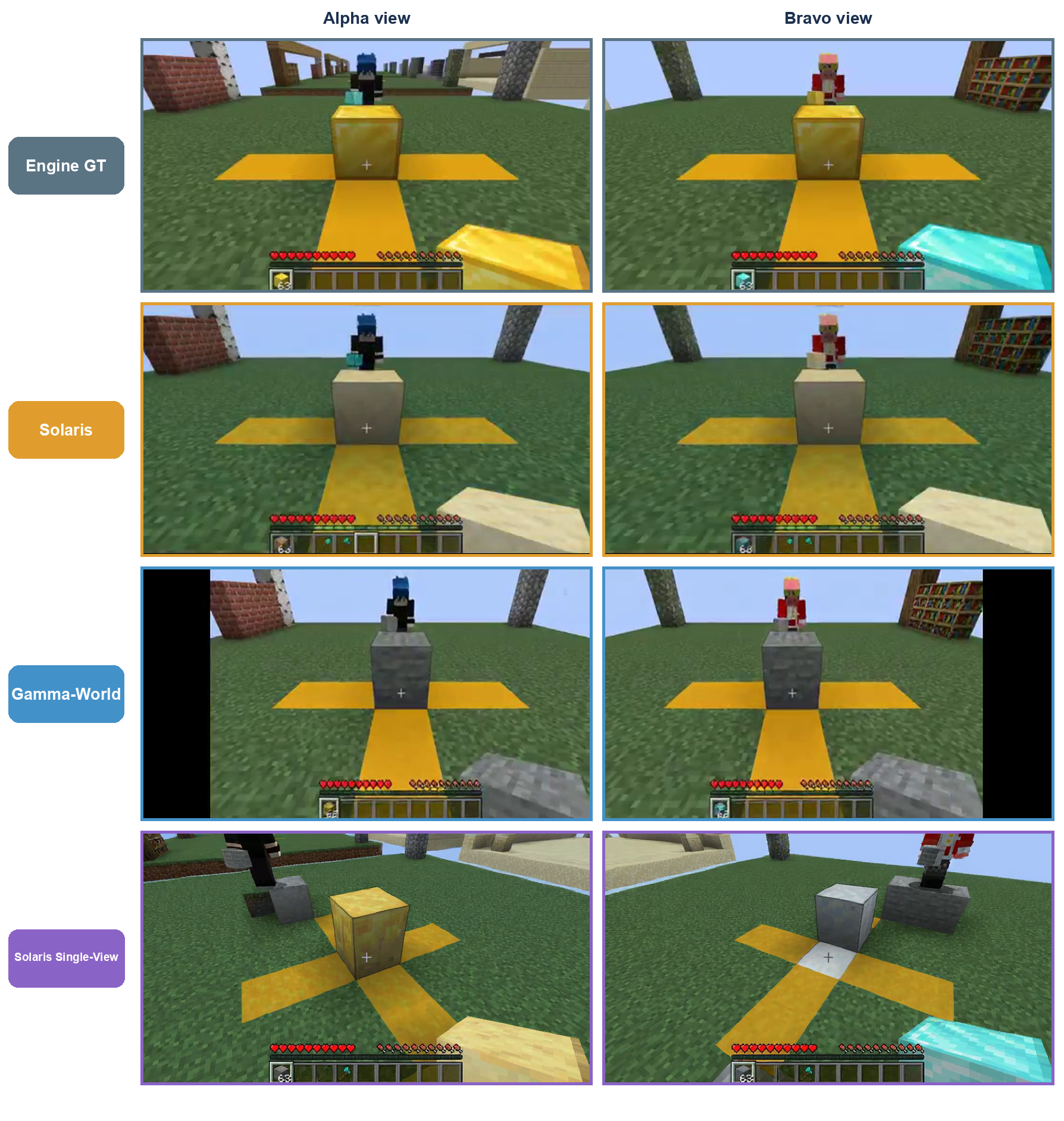}

    \captionof{figure}{
        \textbf{Race-condition consistency}
        (case \texttt{MWB3-SW-RC001}).
        Alpha and Bravo concurrently place gold and diamond blocks at
        the same target. The resulting Alpha and Bravo views are
        shown for Engine GT, Solaris (MP), Gamma-World, and Solaris (SP).
    }
    \label{fig:app-race-condition}
\end{minipage}

\clearpage
\subsection{Delayed Revisit}

\paragraph{Style and ID consistency.}
Figure~\ref{fig:app-revisit-style-id} compares the initial Alpha view
with Bravo's delayed return to the same location.

\noindent
\begin{minipage}{\textwidth}
    \centering
    \includegraphics[
        width=\textwidth,
        height=0.78\textheight,
        keepaspectratio
    ]{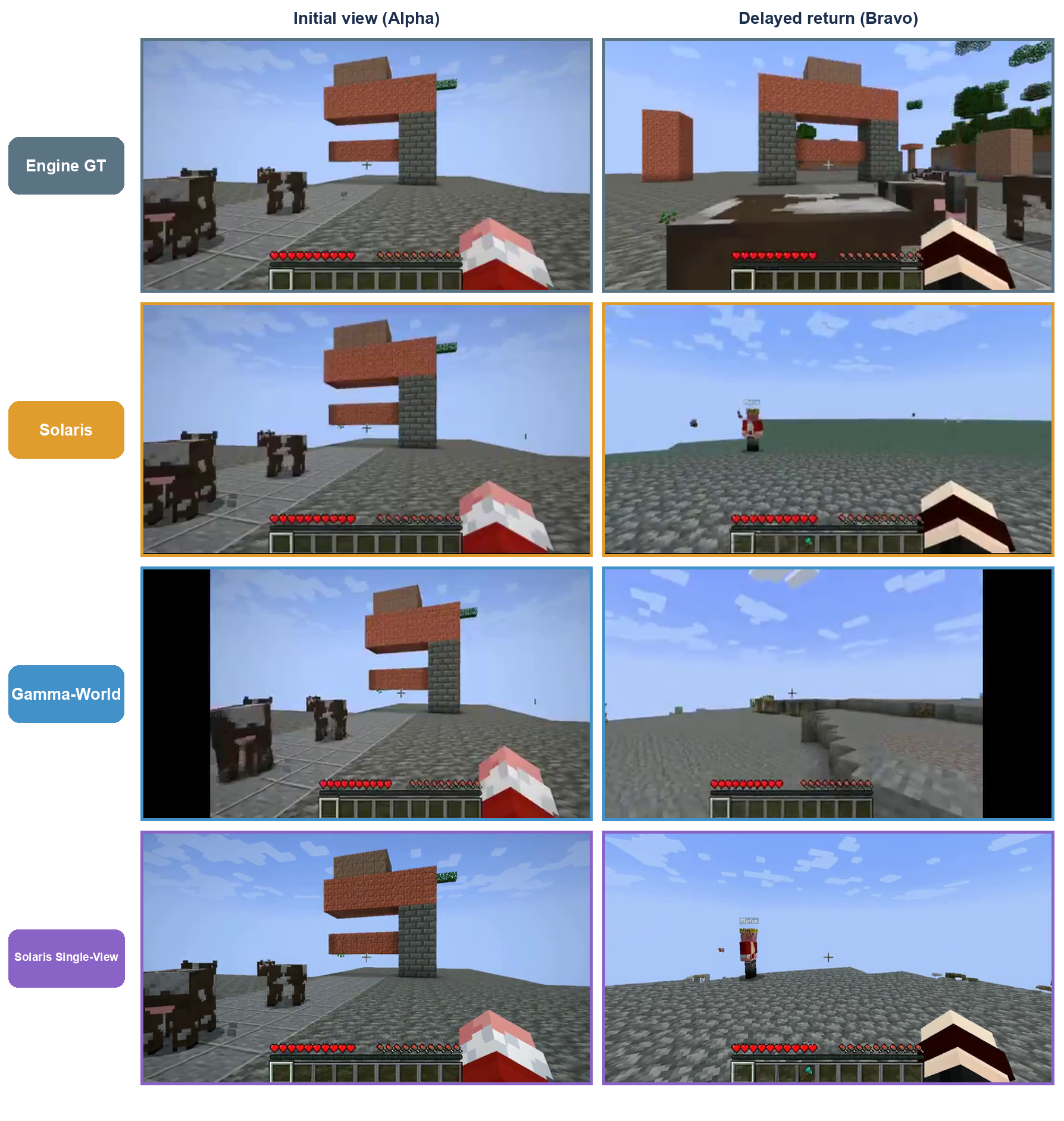}

    \captionof{figure}{
        \textbf{Style and ID consistency under delayed revisit}
        (case \texttt{MWB3-R-H09P}).
        The initial Alpha view and the delayed Bravo view are shown
        for Engine GT, Solaris (MP), Gamma-World, and Solaris (SP).
    }
    \label{fig:app-revisit-style-id}
\end{minipage}

\clearpage
\section{Human Preference Annotation}
\label{app:human-preference-annotation}

This appendix describes the human-preference annotation procedure,
including task construction, the annotation interface, annotator
training, and quality-control measures.

\subsection{Annotation Procedure}
\label{app:human-annotation-procedure}

\paragraph{Task construction.}
The human study covers ten evaluation dimensions. For each dimension,
we sample five benchmark cases and conduct complete pairwise comparisons
among four systems: Engine GT, Solaris (MP), Gamma-World, and Solaris (SP).
Each case therefore contains
\begin{equation}
\binom{4}{2}=6
\end{equation}
system pairs, resulting in
\begin{equation}
10\times5\times\binom{4}{2}=300
\end{equation}
pairwise comparisons in the main annotation study. Each comparison is
independently assessed by three annotators. Before entering the main
study, each annotator completes 30 pre-annotation trials. Each annotator
therefore completes 330 tasks in total, including the pre-annotation
trials and the main study.

\paragraph{Dimension-specific judgments.}
Each annotation task evaluates only one capability dimension. The
annotators are instructed to judge the two system outputs according to
the dimension-specific criterion and to ignore unrelated differences.
For example, building identity consistency focuses on whether the
distinctive layout, silhouette, openings, roof, and other recognizable
components of the same building are preserved after revisiting. General
image quality is ignored when it does not affect the identity judgment.
The same video pair may therefore receive different judgments under
different evaluation dimensions.

\paragraph{Annotation interface.}
Figure~\ref{fig:human-annotation-interface} shows the annotation
interface. The top section presents the evaluation dimension and its
corresponding question. A case-specific evaluation guide describes the
action sequence, the expected behavior, and the major failure patterns.
Separate \textsc{Focus} and \textsc{Ignore} instructions further clarify
which visual evidence should and should not affect the judgment.

The middle section displays the outputs as System A and System B,
without revealing their underlying system identities. Each system panel
contains the synchronized Alpha and Bravo player views. Annotators can
replay or play both system outputs together, pause playback, and adjust
the playback speed. After viewing the two outputs, the annotator selects
one of four responses: \emph{A Better}, \emph{Tie}, \emph{B Better}, or
\emph{Cannot Judge}. The final option is used when missing content,
playback failure, or insufficient visual evidence prevents a reliable
comparison. Responses are saved locally and exported after the
annotation session.

\noindent
\begin{minipage}{\linewidth}
\centering
\includegraphics[
    width=\linewidth,
    height=0.86\textheight,
    keepaspectratio
]{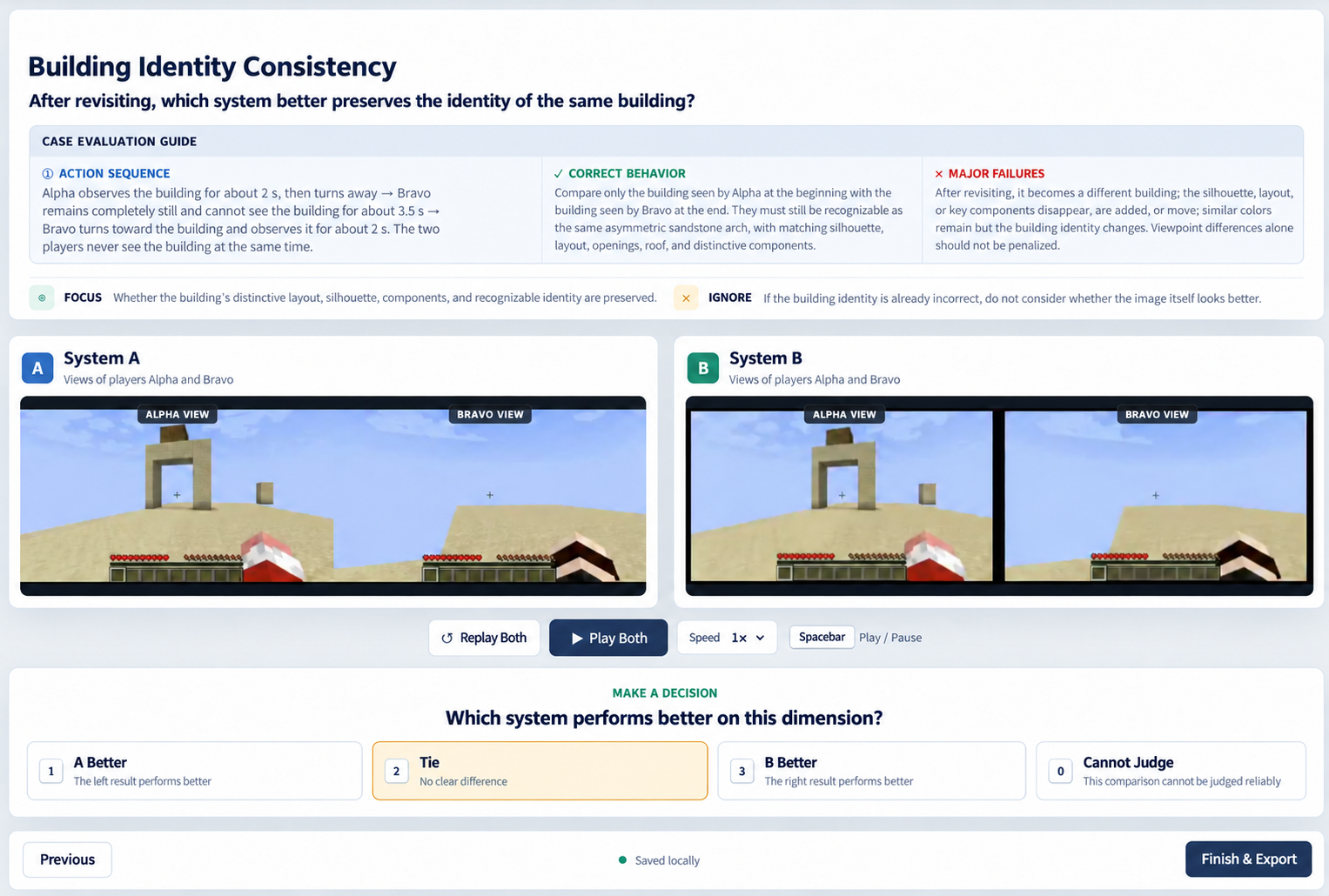}
\captionof{figure}{
    \textbf{Human-preference annotation interface.}
    The interface provides a dimension-specific question, a case-level
    evaluation guide, explicit focus and ignore criteria, synchronized
    Alpha and Bravo views for Systems A and B, joint playback controls,
    and four response options: A Better, Tie, B Better, and Cannot Judge.
}
\label{fig:human-annotation-interface}
\end{minipage}

\subsection{Annotator Training and Quality Control}
\label{app:human-annotation-quality-control}

\paragraph{Pre-annotation test.}
Before the main study, each annotator completes 30 pre-annotation
samples covering the evaluation dimensions and the four response types.
The pre-annotation phase familiarizes annotators with the interface,
dimension-specific definitions, expected behaviors, and common failure
patterns. Incorrect or ambiguous judgments are reviewed with the
annotators before they proceed to the main annotation tasks.

\paragraph{Random inspection and relabeling.}
We randomly inspect 20\% of the completed annotations in each annotation
batch. The inspection checks whether the selected response follows the
dimension-specific instructions and whether the available video evidence
supports the judgment. If the error rate of the inspected annotations in
a batch exceeds 10\%, the entire batch is reassigned to a different
annotator for independent relabeling. This procedure is applied
throughout the study to identify systematic misunderstandings and reduce
residual annotation errors.

\end{document}